\documentclass[journal]{journal}
\usepackage[utf8]{inputenc}
\usepackage[T1]{fontenc}
\usepackage{amsmath}
\usepackage{graphicx}
\usepackage{float}

\usepackage{tikz}
\usepackage{mathdots}
\usepackage{yhmath}
\usepackage{cancel}
\usepackage{color}
\usepackage{siunitx}
\usepackage{array}
\usepackage{multirow}
\usepackage{amssymb}
\usepackage{gensymb}
\usepackage{tabularx}
\usepackage{booktabs}
\usetikzlibrary{fadings}
\usetikzlibrary{patterns}
\usetikzlibrary{shadows.blur}

\usepackage{caption}
\usepackage{subcaption}
\usepackage[hyphens]{url}
\usepackage{makecell}

\usepackage{tabularx}
\usepackage{options}
\let\labelindent\relax
\usepackage{enumitem}
\usepackage{times,mathtools}
\usepackage{microtype,xparse,tcolorbox}
\usepackage{footnote}
\usepackage{ragged2e}
\usepackage{authblk}
\allowdisplaybreaks
\usepackage{fancyhdr}

\ifCLASSINFOpdf
\else
\fi

\begin{document}
%
\title{Facial Expression Phoenix (FePh): An Annotated Sequenced Dataset for Facial and Emotion-Specified Expressions in Sign Language}
%
%
%

\author{Marie~Alaghband,
        Niloofar~Yousefi,
        and~Ivan~Garibay
\thanks{M. Alaghband, N. Yousefi, and I. Garibay are with the Department of Industrial Engineering and Management Systems, University of Central Florida, Orlando, Florida, USA. E-mail: \texttt{marie.alaghband@knights.ucf.edu}, \texttt{niloofar.yousefi@ucf.edu}, and 
\texttt{igaribay@ucf.edu}.}
}

%

\maketitle
\thispagestyle{empty}

\begin{abstract}
Facial expressions are important parts of both gesture and sign language recognition systems. Despite the recent advances in both fields, annotated facial expression datasets in the context of sign language are still scarce resources. In this manuscript, we introduce an annotated sequenced facial expression dataset in the context of  sign language, comprising over $3000$ facial images extracted from the daily news and weather forecast of the public tv-station PHOENIX. Unlike the majority of currently existing facial expression datasets, FePh provides sequenced semi-blurry facial images with different head poses, orientations, and movements. In addition, in the majority of images, identities are mouthing the words, which makes the data more challenging. To annotate this dataset we consider primary, secondary, and tertiary dyads of seven basic emotions of "sad", "surprise", "fear", "angry", "neutral", "disgust", and "happy". We also considered the "None" class if the image's facial expression could not be described by any of the aforementioned emotions. Although we provide FePh as a facial expression dataset of signers in sign language, it has a wider application in gesture recognition and Human Computer Interaction (HCI) systems.\\
\end{abstract}


\ifCLASSOPTIONpeerreview
\begin{center} \bfseries EDICS Category: 3-BBND \end{center}
\fi
%
\IEEEpeerreviewmaketitle

\section{Introduction}
\IEEEPARstart{A}{bout} $466$ million people worldwide have disabled hearing and by $2050$ this number will increase to $900$ million people--one in every ten (based on the most recent report of World health Organization \cite{cite1} (WHO) on January 2020). People with hearing loss often communicate through lip-reading skills, texts, and sign language. Sign language, the natural language of people with severe or profound hearing loss, is unique to each country or region \cite{cite2} and has its own grammar and structure. Combinations of hand movements, postures, and shapes with facial expressions make sign language very unique and complex \cite{cite3}.

Hands are widely used in sign language to convey meaning. Despite the importance of hands in sign language, the directions of eye gazes, eyebrows, eye blinks, and mouths, as part of facial expressions also play an integral role in conveying both emotion and grammar \cite{cite4}. Facial expressions that support grammatical constructions in sign language help to eliminate the ambiguity of signs \cite{cite5}. Therefore, sign language recognition systems without facial expression recognition are incomplete \cite{cite6}.

Due to the importance of hand shapes and movements in sign language, hands are widely used in sign language recognition systems. However, the other integral part of sign language, facial expressions, has not yet been well studied. One reason is the lack of an annotated facial expression dataset in the context of sign language. To the best of our knowledge, an annotated vision-based facial expression dataset in the field of sign language is a very rare resource. This limits researchers' ability to study multi-modal sign language recognition models that consider both facial expressions and hand gestures.

Conducted research in sign language recognition systems can be categorized in two main groups: vision-based and hardware-based recognition systems. Hardware-based recognition systems use datasets that are collected utilizing special colored gloves \cite{cite9, cite10, cite11}, special sensors, and/or depth cameras (such as Microsoft Kinect and Leap Motion) \cite{cite5, cite12, cite13, cite14, cite15, cite16, cite17, cite42} to capture special features of the signer's gestures. Some well known hardware-based datasets are listed in Table \ref{hardware-based}.

Although utilizing hardware eases the process of capturing special features, they limit the applicability where such hardwares are not available. Therefore, vision-based sign language recognition systems utilizing datasets collected by regular cameras are proposed \cite{cite2, cite7, cite18, cite19, cite20}.

One well known continuous sign language dataset is the RWTH-PHOENIX-Weather corpus \cite{cite18}. RWTH-PHOENIX-Weather is a large vocabulary (more than $1200$ signs) image corpus containing weather forecasts recorded from German news. Two years later, its publicly available extension called RWTH-PHOENIX-Weather multisigner 2014 dataset was introduced. We will create our annotated facial expression dataset based upon RWTH-PHOENIX-Weather multisigner 2014 continuous sign language benchmark dataset.

Facial expression recognition is a very well established field of research with publicly available databases containing basic universal expressions. CK$^+$ 
\cite{cite36} is a well known facial expression database with $327$ annotated video sequences on seven basic universal facial expressions ("anger", "contempt", "disgust", "fear", "happiness", "sadness", and "surprise"). Some other widely used facial expression databases are MMI
\cite{cite37, cite38}, Oulu-CASIA 
\cite{cite39}, and FER2013
\cite{cite40}. FER2013 is an unconstrained large database considering seven emotions (previous six emotions plus "neutral"). Despite the value of these facial expression databases, hardly any one them are in the context of sign language. 

In this paper, we introduce FePh, an annotated facial expression dataset for the publicly available continuous sign language dataset RWTH-PHOENIX-Weather 2014 \cite{cite7}.

As a matter of continuity image data, FePh is similar to CK+, MMI, and Oulu-CASIA. However, it is more complex than those databases as it contains real-life captured videos with more than one basic facial expression in each sequence, with different head poses, orientations, and movements. 

\begin{table}
\caption{An overview of some hardware-based sign language datasets}
\label{hardware-based}
\resizebox{0.5\textwidth}{!}{%
\begin{tabular}{p{0.6cm}p{2.1cm}p{0.8cm}p{.8cm}p{.9cm}p{0.5cm}p{.8cm}}
\hline
Authors & Name & Language & Gesture Type & Language level &  Classes & Data Type\\
\hline

\cite{cite21} & UCI Australian Auslan Sign Language dataset
& Australian & Dynamic & Alphabets & 95 & Data Glove \\[12pt]

\cite{cite12} & ASL Finger Spelling A 
& American & Static & Alphabets & 24 & Depth Images \\[10pt]

\cite{cite12} & ASL Finger Spelling B 
& American & Static & Alphabets & 24 & Depth Image\\[10pt]

\cite{cite22} & MSRGesture $3$D 
& American & - & Words & 12 & Depth Video\\[10pt]

\cite{cite23} & CLAP14 & Italian & - & Words & 20 & Depth Video \\[10pt]

\cite{cite24} & ChaLearn LAP IsoGD
\& ConGD
& - & Static \& Dynamic & - & 249 & RGB \& Depth Video \\[10pt]

\cite{cite25} & PSL Kinect 30
& Polish & Dynamic & Words & 30 & Kinect Video\\[10pt]


 
\cite{cite14} & ISL
& Indian & static & Alphabets, Numbers, Words & 140 & Depth Images \\
\end{tabular}%
}
\end{table}

\begin{table}
\caption{An overview of some vision-based sign language datasets}
\label{hardware-based}
\resizebox{0.5\textwidth}{!}{%
\begin{tabular}{p{0.6cm}p{2.1cm}p{0.8cm}p{.8cm}p{.9cm}p{0.5cm}p{.8cm}}
\hline
Authors & Name & Language & Gesture Type & Language level &  Classes & Data Type\\
\hline

\cite{cite29} & - & American & Static & Alphabets, numbers & 36 & Image \\

\cite{cite12} & ASL Finger Spelling A 
& American & Static & Alphabets & 24 & Image \\[10pt]

\cite{cite30} & HUST-ASL
& American & Static & Alphabets, Numbers & 34 & RGB \& Kinect Image\\[10pt]

\cite{cite19, cite26} & Purdue RVL-SLLL ASL Database
& American & - & Alphabets, Numbers, Words, \hfill \break Paragraphs & 104 & Image, Video \\[25pt]

\cite{cite31} & Boston ASLLVD
& American & Dynamic & Words & \textgreater{}3300 & Video\\[10pt]

\cite{cite20} & ASL-LEX
& American & - & Words & Nearly 1000 & Video\\[10pt]

\cite{cite2} & MS-ASL 
& American & Dynamic & - &  1000 & Video \\[10pt]

\cite{cite32} & - & Arabic & - & Words & 23 & Video \\[10pt]

\cite{cite18} & RWTH-PHOENIX-Weather 2012
& German & - & Sentence & 1200 & Image \\[15pt]

\cite{cite7, cite33} & RWTH-PHOENIX-Weather Multisigner 2014
& German & Dynamic & Sentence & 1080 & Video \\[25pt]

\cite{cite43} & RWTH-PHOENIX-Weather 2014 T
& German & Dynamic & Sentence & 1066 & Video \\[25pt]

\cite{cite27} & SIGNUM
& German & - & words, \hfill \break Sentences & 450 Words, 780 Sentence & Video \\[10pt]

\cite{cite10} & LSA16
& Argentinian & - & Alphabets, Words & 16 & Image \\[10pt]

\cite{cite10} & LSA64
& Argentinian & - & Words & 64 & Video \\[10pt]

\cite{cite14} & the ISL dataset
& Indian & static & Alphabets, Numbers, Words & 140 & Image \\[10pt]

\cite{cite34} & ISL hand shape dataset
& Irish & Static \& Dynamic & - & 23 Static \& 3 Dynamic & Image Video \\[10pt]

\cite{cite35} & Japaneese Finger spelling sign language dataset & Japan & - & - & 41 & Image \\[13pt]
\end{tabular}
}
\end{table}

In addition, FePh not only contains seven basic facial expressions of the FER2013 database, but it also considers their primary, secondary, and tertiary dyads. It is also noteworthy to mention that in sign language, signers mouth the words or sentences to help their audience better grasp meanings. This is a characteristic that makes facial expression datasets in the context of sign language more challenging. As such, this manuscript provides the following contributions: first, introducing annotated facial expression dataset of the RWTH-PHOENIX-Weather 2014 dataset. Second, attributing highly used hand shapes with their associated performed facial expressions, and third, illustrating the relationships between hand shapes and facial expressions in sign language.

\section{Methods}
\IEEEPARstart{D}{ue} to the integral role of facial expressions in conveying emotions and grammar in sign language, it is important to use multi-modal sign language recognition models that consider both hand shapes and facial expressions. Therefore, to create FePh and annotate facial expressions of a sign language dataset with annotated hand shapes, we considered the well-known publicly available continuous RWTH-PHOENIX-Weather 2014 dataset. Since the annotated hand shapes dataset RWTH-PHOENIX-Weather 2014 is publicly available as RWTH-PHOENIX-Weather 2014 MS Handshapes dataset \cite{cite8}, we provide facial expression annotations for the same dataset, which enables researchers to utilize a dataset that has both hand shape and facial expression annotations.

As a starting point, we collected the full frame images of RWTH-PHOENIX-Weather 2014 development set that are identical to the RWTH-PHOENIX-Weather 2014 MS Handshapes dataset \cite{cite8}. Furthermore, in order to create a solid facial expression dataset and avoid the influence of hand shapes on the facial expression annotators, faces of all full frame images are automatically detected, tracked, and cropped using facial recognition techniques.
 
Twelve annotators (six women and six men) between 20 to 40 years old were asked to annotate the data. We asked annotators to answer three questions about each static image: the signer's emotion, visibility, and gender. In terms of emotion, annotators could choose one or more of the following applicable basic universal facial expressions for each static image: "sad", "surprise", "fear", "angry", "neutral", "disgust", and "happy". Although more than seven emotions and their primary, secondary, and tertiary dyads exist, considering all of them was not within the scale of this project. Therefore, we offered the eighth class of "None" as well. Annotators were asked to choose the "None" class when none of the aforementioned emotions could describe the facial expression of the image. In addition, since annotators could choose more than one facial expression for each individual image, the combinations of basic universal facial expressions were also considered (interestingly, this did not result in choosing more than two emotions for each image) and shown by a "\_" in between such as surprise\_fear. The sequence of emotions is not important in the secondary and tertiary dyads (i.e., surprise\_fear and fear\_surprise are the same).

\begin{figure}[t]
\begin{center}
    \includegraphics[width=\textwidth,height=2cm,width=8cm, scale=0.6]{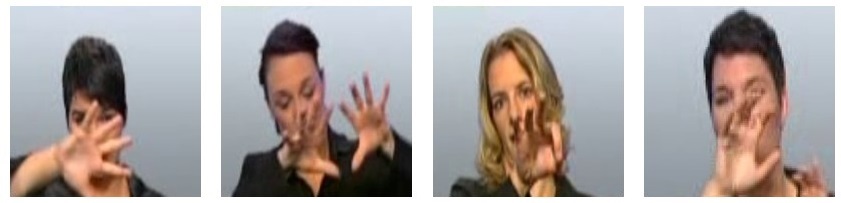}
    \caption{Exemplary images of obscured faces}
    \label{face_not_visible}
\end{center}
\end{figure}

With regard to the second question, visibility, we asked the annotators to evaluate whether the signer's face is completely visible. Although the signer's face was visible in majority of images, this was not always the case. The partial visibility of the face was due to the signer's head movement, position, hand movement, and transitions from one emotion to another emotion. This helped us to detect and opt out these obscured images in the data. Figure \ref{face_not_visible} shows some obscured exemplary images.

The last question of signer's gender was asked to provide statistics of signers' gender. This statistics enables conducting future researches in the affects of gender in expressing emotions and facial expressions.

For our labelling purpose, we took advantage of the Labelbox \cite{cite41} annotating solution tool through which we defined an annotation project and randomly distributed images to be labeled by the annotators. In addition, due to the complexities of the facial images of the RWTH-PHOENIX-Weather 2014 dataset, we used the auto consensus option of the Labelbox tool. These complexities are as follows:
\begin{itemize}
    \item The ambiguity of images, due to signer’s movement, head position, and transitions from one emotion to another (e.g., eyes are closed and/or the lips are still open).
    \item Low quality (resolution) and blurriness of images.
    \item Mouthed words that confuse facial expression annotators.
    \item Personal differences between signers expressing facial expressions.
    \item The best facial expression that describes the image is not included in the dataset.
    \item Images may not be in facial expression's top frame.
    \item Large intra-class variance (such as "surprised" emotion with open or closed mouth).
    \item Inter-class similarities.
\end{itemize}

With the usage of auto consensus option of Labelbox, we asked more than one annotator (three) 
to annotate about $60\%$ percent of the data. For the images with three labels, we chose the most voted emotion as the final label of the facial image. In cases where there was not a most voted emotion, but the image was a part of a sequence of images, we have assigned labels based on the before or after images' facial expression of the same sequence. On the other hand, if there was not a most voted emotion, and the image was not a part of a sequence of images (i.e., one single image without any sequence), we asked our annotators to relabel the image. In this case, all images needed to be labelled by three different annotators.

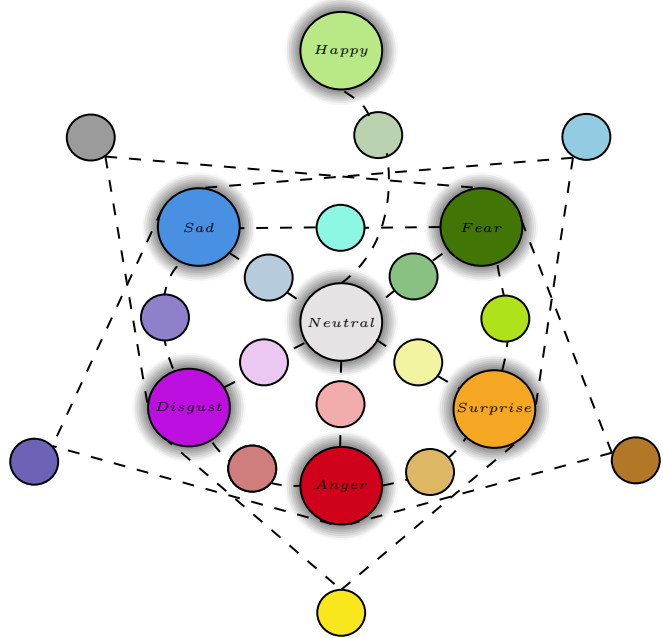
\begin{figure*}[t]
    \begin{center}
        \tikzset{every picture/.style={line width=0.75pt}} 
        \begin{tikzpicture}[x=0.75pt,y=0.75pt,yscale=-1.1,xscale=1.1]
        
        \draw  [fill={rgb, 255:red, 74; green, 243; blue, 211 }  ,fill opacity=0.62 ] (190.44,100.19) .. controls (190.44,94.36) and (195.36,89.64) .. (201.44,89.64) .. controls (207.51,89.64) and (212.43,94.36) .. (212.43,100.19) .. controls (212.43,106.02) and (207.51,110.74) .. (201.44,110.74) .. controls (195.36,110.74) and (190.44,106.02) .. (190.44,100.19) -- cycle ;
        \draw  [fill={rgb, 255:red, 207; green, 125; blue, 125 }  ,fill opacity=1 ] (149.95,210.68) .. controls (149.95,204.85) and (154.87,200.12) .. (160.94,200.12) .. controls (167.01,200.12) and (171.93,204.85) .. (171.93,210.68) .. controls (171.93,216.5) and (167.01,221.23) .. (160.94,221.23) .. controls (154.87,221.23) and (149.95,216.5) .. (149.95,210.68) -- cycle ;
        \draw  [fill={rgb, 255:red, 229; green, 227; blue, 227 }  ,fill opacity=1 ][blur shadow={shadow xshift=0pt,shadow yshift=0pt, shadow blur radius=6pt, shadow blur steps=8 ,shadow opacity=100}] (183.04,143.4) .. controls (183.04,133.57) and (191.43,125.61) .. (201.78,125.61) .. controls (212.13,125.61) and (220.52,133.57) .. (220.52,143.4) .. controls (220.52,153.23) and (212.13,161.2) .. (201.78,161.2) .. controls (191.43,161.2) and (183.04,153.23) .. (183.04,143.4) -- cycle ;
        \draw  [fill={rgb, 255:red, 248; green, 231; blue, 28 }  ,fill opacity=1 ] (190.75,276.84) .. controls (190.75,271.01) and (195.67,266.28) .. (201.74,266.28) .. controls (207.81,266.28) and (212.73,271.01) .. (212.73,276.84) .. controls (212.73,282.66) and (207.81,287.39) .. (201.74,287.39) .. controls (195.67,287.39) and (190.75,282.66) .. (190.75,276.84) -- cycle ;
        \draw  [fill={rgb, 255:red, 155; green, 155; blue, 155 }  ,fill opacity=1 ] (75.92,58.76) .. controls (75.92,52.93) and (80.84,48.21) .. (86.91,48.21) .. controls (92.98,48.21) and (97.9,52.93) .. (97.9,58.76) .. controls (97.9,64.58) and (92.98,69.31) .. (86.91,69.31) .. controls (80.84,69.31) and (75.92,64.58) .. (75.92,58.76) -- cycle ;
        \draw  [fill={rgb, 255:red, 108; green, 99; blue, 183 }  ,fill opacity=1 ] (49.99,207.51) .. controls (49.99,201.68) and (54.91,196.96) .. (60.98,196.96) .. controls (67.06,196.96) and (71.98,201.68) .. (71.98,207.51) .. controls (71.98,213.34) and (67.06,218.06) .. (60.98,218.06) .. controls (54.91,218.06) and (49.99,213.34) .. (49.99,207.51) -- cycle ;
        \draw  [fill={rgb, 255:red, 178; green, 120; blue, 40 }  ,fill opacity=1 ] (325.81,206.89) .. controls (325.81,201.06) and (330.73,196.34) .. (336.8,196.34) .. controls (342.87,196.34) and (347.79,201.06) .. (347.79,206.89) .. controls (347.79,212.71) and (342.87,217.44) .. (336.8,217.44) .. controls (330.73,217.44) and (325.81,212.71) .. (325.81,206.89) -- cycle ;
        \draw  [fill={rgb, 255:red, 147; green, 204; blue, 226 }  ,fill opacity=1 ] (303.15,58.47) .. controls (303.15,52.64) and (308.07,47.92) .. (314.15,47.92) .. controls (320.22,47.92) and (325.14,52.64) .. (325.14,58.47) .. controls (325.14,64.29) and (320.22,69.02) .. (314.15,69.02) .. controls (308.07,69.02) and (303.15,64.29) .. (303.15,58.47) -- cycle ;
        \draw  [fill={rgb, 255:red, 189; green, 16; blue, 224 }  ,fill opacity=1 ][blur shadow={shadow xshift=0pt,shadow yshift=0pt, shadow blur radius=6pt, shadow blur steps=8 ,shadow opacity=100}] (113.2,182.66) .. controls (113.2,172.84) and (121.59,164.87) .. (131.94,164.87) .. controls (142.29,164.87) and (150.68,172.84) .. (150.68,182.66) .. controls (150.68,192.49) and (142.29,200.46) .. (131.94,200.46) .. controls (121.59,200.46) and (113.2,192.49) .. (113.2,182.66) -- cycle ;
        \draw  [fill={rgb, 255:red, 74; green, 144; blue, 226 }  ,fill opacity=1 ][blur shadow={shadow xshift=0pt,shadow yshift=0pt, shadow blur radius=6pt, shadow blur steps=8 ,shadow opacity=100}] (117.7,99.83) .. controls (117.7,90) and (126.09,82.03) .. (136.44,82.03) .. controls (146.79,82.03) and (155.18,90) .. (155.18,99.83) .. controls (155.18,109.66) and (146.79,117.62) .. (136.44,117.62) .. controls (126.09,117.62) and (117.7,109.66) .. (117.7,99.83) -- cycle ;
        \draw  [fill={rgb, 255:red, 65; green, 117; blue, 5 }  ,fill opacity=1 ][blur shadow={shadow xshift=0pt,shadow yshift=0pt, shadow blur radius=6pt, shadow blur steps=8 ,shadow opacity=100}] (247.29,99.83) .. controls (247.29,90) and (255.68,82.03) .. (266.03,82.03) .. controls (276.38,82.03) and (284.77,90) .. (284.77,99.83) .. controls (284.77,109.66) and (276.38,117.62) .. (266.03,117.62) .. controls (255.68,117.62) and (247.29,109.66) .. (247.29,99.83) -- cycle ;
        \draw  [color={rgb, 255:red, 0; green, 0; blue, 0 }  ,draw opacity=1 ][fill={rgb, 255:red, 184; green, 233; blue, 134 }  ,fill opacity=1 ][blur shadow={shadow xshift=0pt,shadow yshift=0pt, shadow blur radius=6pt, shadow blur steps=8 ,shadow opacity=100}] (183.04,18.91) .. controls (183.04,9.08) and (191.43,1.11) .. (201.78,1.11) .. controls (212.13,1.11) and (220.52,9.08) .. (220.52,18.91) .. controls (220.52,28.74) and (212.13,36.7) .. (201.78,36.7) .. controls (191.43,36.7) and (183.04,28.74) .. (183.04,18.91) -- cycle ;
        \draw  [fill={rgb, 255:red, 245; green, 166; blue, 35 }  ,fill opacity=1 ][blur shadow={shadow xshift=0pt,shadow yshift=0pt, shadow blur radius=6pt, shadow blur steps=8 ,shadow opacity=100}] (253.23,183.34) .. controls (253.23,173.51) and (261.62,165.55) .. (271.97,165.55) .. controls (282.32,165.55) and (290.71,173.51) .. (290.71,183.34) .. controls (290.71,193.17) and (282.32,201.14) .. (271.97,201.14) .. controls (261.62,201.14) and (253.23,193.17) .. (253.23,183.34) -- cycle ;
        \draw  [fill={rgb, 255:red, 208; green, 2; blue, 27 }  ,fill opacity=1 ][blur shadow={shadow xshift=0pt,shadow yshift=0pt, shadow blur radius=6pt, shadow blur steps=8 ,shadow opacity=100}] (183.04,218.51) .. controls (183.04,208.69) and (191.43,200.72) .. (201.78,200.72) .. controls (212.13,200.72) and (220.52,208.69) .. (220.52,218.51) .. controls (220.52,228.34) and (212.13,236.31) .. (201.78,236.31) .. controls (191.43,236.31) and (183.04,228.34) .. (183.04,218.51) -- cycle ;
        \draw  [fill={rgb, 255:red, 226; green, 176; blue, 237 }  ,fill opacity=0.69 ] (155.35,161.92) .. controls (155.35,156.09) and (160.27,151.36) .. (166.34,151.36) .. controls (172.41,151.36) and (177.33,156.09) .. (177.33,161.92) .. controls (177.33,167.74) and (172.41,172.47) .. (166.34,172.47) .. controls (160.27,172.47) and (155.35,167.74) .. (155.35,161.92) -- cycle ;
        \draw  [fill={rgb, 255:red, 136; green, 193; blue, 130 }  ,fill opacity=1 ] (223.92,122.49) .. controls (223.92,116.67) and (228.84,111.94) .. (234.91,111.94) .. controls (240.98,111.94) and (245.91,116.67) .. (245.91,122.49) .. controls (245.91,128.32) and (240.98,133.04) .. (234.91,133.04) .. controls (228.84,133.04) and (223.92,128.32) .. (223.92,122.49) -- cycle ;
        \draw  [fill={rgb, 255:red, 139; green, 173; blue, 200 }  ,fill opacity=0.63 ] (157.51,123.01) .. controls (157.51,117.18) and (162.43,112.46) .. (168.5,112.46) .. controls (174.57,112.46) and (179.49,117.18) .. (179.49,123.01) .. controls (179.49,128.84) and (174.57,133.56) .. (168.5,133.56) .. controls (162.43,133.56) and (157.51,128.84) .. (157.51,123.01) -- cycle ;
        \draw  [fill={rgb, 255:red, 243; green, 172; blue, 172 }  ,fill opacity=1 ] (190.44,181.11) .. controls (190.44,175.28) and (195.36,170.56) .. (201.44,170.56) .. controls (207.51,170.56) and (212.43,175.28) .. (212.43,181.11) .. controls (212.43,186.94) and (207.51,191.66) .. (201.44,191.66) .. controls (195.36,191.66) and (190.44,186.94) .. (190.44,181.11) -- cycle ;
        \draw  [fill={rgb, 255:red, 243; green, 244; blue, 161 }  ,fill opacity=1 ] (226.08,161.92) .. controls (226.08,156.09) and (231,151.36) .. (237.07,151.36) .. controls (243.14,151.36) and (248.07,156.09) .. (248.07,161.92) .. controls (248.07,167.74) and (243.14,172.47) .. (237.07,172.47) .. controls (231,172.47) and (226.08,167.74) .. (226.08,161.92) -- cycle ;
        \draw  [fill={rgb, 255:red, 186; green, 210; blue, 175 }  ,fill opacity=1 ] (207.72,57.65) .. controls (207.72,51.83) and (212.64,47.1) .. (218.71,47.1) .. controls (224.78,47.1) and (229.71,51.83) .. (229.71,57.65) .. controls (229.71,63.48) and (224.78,68.2) .. (218.71,68.2) .. controls (212.64,68.2) and (207.72,63.48) .. (207.72,57.65) -- cycle ;
        \draw  [dash pattern={on 4.5pt off 4.5pt}]  (93.62,67.51) -- (113.2,182.66) ;
        \draw  [dash pattern={on 4.5pt off 4.5pt}]  (197.3,236.09) -- (69.32,200.3) ;
        \draw  [dash pattern={on 4.5pt off 4.5pt}]  (307.99,68.03) -- (136.44,82.03) ;
        \draw  [dash pattern={on 4.5pt off 4.5pt}]  (307.99,68.03) -- (290.71,183.34) ;
        \draw  [dash pattern={on 4.5pt off 4.5pt}]  (119.54,91.89) -- (69.32,200.3) ;
        \draw  [dash pattern={on 4.5pt off 4.5pt}]  (93.62,67.51) -- (266.03,82.03) ;
        \draw  [dash pattern={on 4.5pt off 4.5pt}]  (325.81,203.41) -- (205.94,236.09) ;
        \draw  [dash pattern={on 4.5pt off 4.5pt}]  (201.74,266.28) -- (282.61,198.23) ;
        \draw  [dash pattern={on 4.5pt off 4.5pt}]  (120.08,196.15) -- (201.74,266.28) ;
        \draw  [dash pattern={on 4.5pt off 4.5pt}]  (325.81,203.41) -- (283.69,95) ;
        \draw  [dash pattern={on 4.5pt off 4.5pt}]  (150.32,111.6) -- (158.96,117.31) ;
        \draw  [dash pattern={on 4.5pt off 4.5pt}]  (226.46,128.72) -- (218.36,135.46) ;
        \draw  [dash pattern={on 4.5pt off 4.5pt}]  (243.74,115.23) -- (250.22,110.56) ;
        \draw  [dash pattern={on 4.5pt off 4.5pt}]  (247.29,99.83) -- (212.43,100.19) ;
        \draw  [dash pattern={on 4.5pt off 4.5pt}]  (201.24,200.72) -- (201.44,191.66) ;
        \draw  [dash pattern={on 4.5pt off 4.5pt}]  (201.78,161.2) -- (201.44,170.56) ;
        \draw  [dash pattern={on 4.5pt off 4.5pt}]  (184.88,153.1) -- (175.7,157.77) ;
        \draw  [dash pattern={on 4.5pt off 4.5pt}]  (148.16,173.33) -- (157.34,168.66) ;
        \draw  [dash pattern={on 4.5pt off 4.5pt}]  (245.9,168.14) -- (255.62,173.85) ;
        \draw  [dash pattern={on 4.5pt off 4.5pt}]  (217.82,151.54) -- (226.46,157.25) ;
        \draw  [dash pattern={on 4.5pt off 4.5pt}]  (176.78,129.76) -- (185.42,135.46) ;
        \draw  [dash pattern={on 4.5pt off 4.5pt}]  (190.05,99.99) -- (155.18,100.35) ;
        \draw  [dash pattern={on 4.5pt off 4.5pt}]  (201.78,125.61) .. controls (223.38,110.04) and (225.38,83.07) .. (222.68,65.95) ;
        \draw  [dash pattern={on 4.5pt off 4.5pt}]  (214.58,48.84) .. controls (211.88,43.13) and (210.8,42.61) .. (201.78,36.7) ;
        \draw  [dash pattern={on 4.5pt off 4.5pt}]  (275.59,165.55) .. controls (281.53,142.2) and (276.13,134.42) .. (273.43,117.31) ;
        \draw  [fill={rgb, 255:red, 176; green, 226; blue, 28 }  ,fill opacity=1 ] (266,141.85) .. controls (266,136.02) and (270.92,131.29) .. (276.99,131.29) .. controls (283.06,131.29) and (287.98,136.02) .. (287.98,141.85) .. controls (287.98,147.67) and (283.06,152.4) .. (276.99,152.4) .. controls (270.92,152.4) and (266,147.67) .. (266,141.85) -- cycle ;
        \draw  [dash pattern={on 4.5pt off 4.5pt}]  (220.52,218.51) .. controls (232.4,215.86) and (246.98,213.79) .. (258.85,196.67) ;
        \draw  [fill={rgb, 255:red, 222; green, 183; blue, 101 }  ,fill opacity=1 ] (231.48,212.23) .. controls (231.48,206.4) and (236.4,201.68) .. (242.47,201.68) .. controls (248.54,201.68) and (253.46,206.4) .. (253.46,212.23) .. controls (253.46,218.06) and (248.54,222.78) .. (242.47,222.78) .. controls (236.4,222.78) and (231.48,218.06) .. (231.48,212.23) -- cycle ;
        \draw  [dash pattern={on 4.5pt off 4.5pt}]  (142.76,198.74) .. controls (148.16,207.04) and (163.82,220.01) .. (183.04,218.51) ;
        \draw  [fill={rgb, 255:red, 207; green, 125; blue, 125 }  ,fill opacity=1 ] (149.95,210.68) .. controls (149.95,204.85) and (154.87,200.12) .. (160.94,200.12) .. controls (167.01,200.12) and (171.93,204.85) .. (171.93,210.68) .. controls (171.93,216.5) and (167.01,221.23) .. (160.94,221.23) .. controls (154.87,221.23) and (149.95,216.5) .. (149.95,210.68) -- cycle ;
        \draw  [dash pattern={on 4.5pt off 4.5pt}]  (126.56,117.82) .. controls (119,125.61) and (116.84,146.87) .. (124.94,165.55) ;
        \draw  [fill={rgb, 255:red, 144; green, 127; blue, 202 }  ,fill opacity=1 ] (109.95,141.33) .. controls (109.95,135.5) and (114.87,130.78) .. (120.94,130.78) .. controls (127.01,130.78) and (131.93,135.5) .. (131.93,141.33) .. controls (131.93,147.15) and (127.01,151.88) .. (120.94,151.88) .. controls (114.87,151.88) and (109.95,147.15) .. (109.95,141.33) -- cycle ;
        
        \draw (201.78,18.91) node  [font=\tiny]  {$Happy$};
        \draw (136.44,99.83) node  [font=\tiny]  {$Sad$};
        \draw (271.97,183.34) node  [font=\tiny]  {$Surprise$};
        \draw (266.03,99.83) node  [font=\tiny]  {$Fear$};
        \draw (201.78,218.51) node  [font=\tiny]  {$Anger$};
        \draw (131.94,182.66) node  [font=\tiny]  {$Disgust$};
        \draw (201.78,143.4) node  [font=\tiny]  {$Neutral$};
        
        \end{tikzpicture}

        \caption{Graph displaying the primary, secondary, and tertiary dyads on seven basic universal emotions in the FePh dataset. Seven basic universal emotions are shown with colored circles containing their name. In addition, colored circles that are connecting each two basic emotions show the secondary or tertiary dyads of the basic emotions that they are connected to. For example, the orange circle connected to the "Disgust" and "Surprise" emotions, shows the emotion of "Disgust\_Surprise". Emotion "Happy"'s only dyad is with "Neutral" emotion named as "Neutral\_Happy". }
    \label{FE_relations}
    \end{center}
\end{figure*}

\section{Data Records}
\IEEEPARstart{T}{he} FePh facial expression dataset produced with the above method, is stored on Harvard Dataverse (\url{https://dataverse.harvard.edu/dataset.xhtml?persistentId=doi:10.7910/DVN/358QMQ}). All facial images are stored in "FePh\_images.zip". Although the full frame images of the FePh dataset are identical to the RWTH-PHOENIX-Weather 2014 images, images' filename in the FePh\_images.zip are different from the original images. FePh filenames consider both image folder\_name and image\_file\_number. For example, the full frame image of the facial image with filename "\textit{01August\_2011\_Monday\_heute\_defa ult-6.avi\_pid0\_fn000054-0.png}" in the FePh dataset is identical to the full frame image with the directory of "\textit{... / 01August\_2011 \_Monday\_heute\_default-6 / 1 / 01August\_2011 \_Monday\_heute.avi\_pid0\_ fn000054-0.png}" in the RWTH-PHOENIX-Weather 2014 and "\textit{... / 01August\_2011 \_Monday\_heute\_default-6 / 1 /\_.png\_ fn000054-0.png}" image in the RWTH-PHOENIX-Weather 2014 MS Handshapes dataset. This helped us to store all images in one single folder (FePh\_images). The FePh\_labels.csv file contains images' filenames, facial expression labels, and gender labels. To ease data usability, we stored the facial expression labels as codes. Table \ref{label_code} shows the facial expression labels with their corresponding code numbers. In addition, $0$ and $1$ in the gender column represent the male and female genders, respectively.

\begin{table}[t]
\caption{The facial expression labels with their corresponding numbers}
\label{label_code}
\centering
\resizebox{0.5\textwidth}{!}{%
\begin{tabular}{ll|ll|ll}
\hline
label\# & Emotion & label\# & Emotion & label\# & Emotion \\
\hline
0 & neutral & 10 & anger\_neutral & 52 & sad\_disgust \\
1 & anger & 21 & disgust\_anger & 53 & sad\_fear \\
2 & disgust & 31 & anger\_fear & 60 & neutral\_surprise \\
3 & fear & 32 & fear\_disgust & 61 & surprise\_anger \\
4 & happy & 40 & neutral\_happy & 62 & surprise\_disgust \\
5 & sad & 50 & neutral\_sad & 63 & surprise\_fear \\
6 & surprise & 51 & sad\_anger & 65 & surprise\_sad \\
7 & none &  &  &  & \\
\end{tabular}%
}
\end{table}

\footnotesize
\begin{figure*}[t]
\begin{subfigure}{\textwidth}
  \centering
  \includegraphics[width=\textwidth,height=6cm,width=15cm, scale=1]{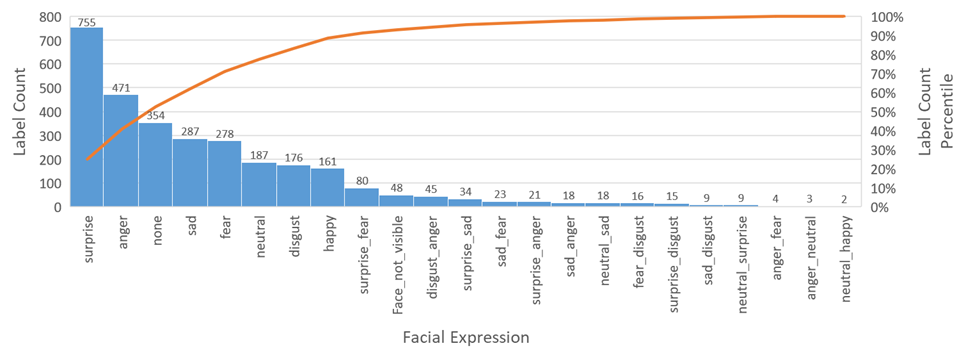}
  \caption{Pareto chart showing the distribution counts per facial expression class in the FePh dataset for the
  top 14 hand shape classes (briefly shown as HSH). As the chart shows, basic facial expressions represent $90\%$ of the data.}
  \label{FE_frequency}
\end{subfigure}

\hspace{2cm}
\medskip

\begin{subfigure}{.45\textwidth}
  \centering
  \includegraphics[width=\linewidth]{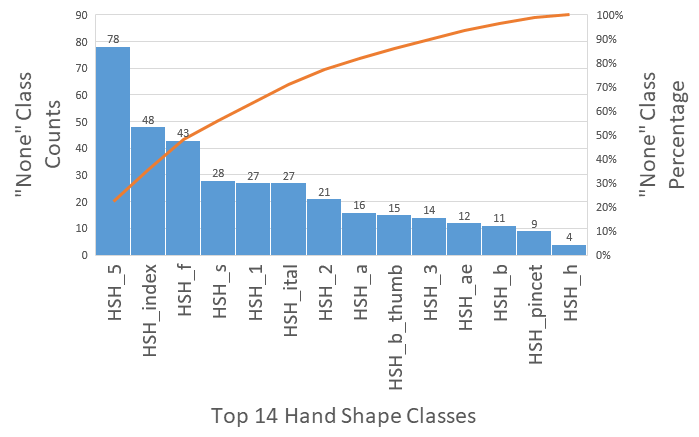}
  \caption{Pareto chart showing the distribution of the "None" facial expression class counts per top 14 hand shape classes (HSH)}
  \label{none_dist}
\end{subfigure}\hspace*{\fill}
\begin{subfigure}{.45\textwidth}
  \centering
  \includegraphics[width=\linewidth]{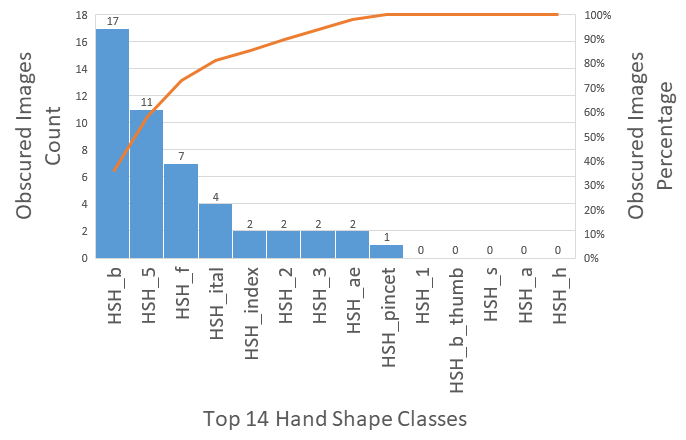}
  \caption{Pareto chart showing the distribution of the obscured image counts per top 14 hand shape classes (HSH)}
  \label{obscured_dist}
\end{subfigure}\hspace*{\fill}
\caption{Pareto chart showing the distribution of the facial expression "None" and obscured image counts for top 14 hand shape classes (HSH).}
\label{dist}
\end{figure*}

\section{Technical Validation}
\normalsize
\IEEEPARstart{T}{he} FePh dataset is created by manually labelling $3359$ images of the RWTH-PHOENIX-Weather 2014 development set that are identical to the full frame images of the RWTH-PHOENIX-Weather 2014 MS Handshapes dataset. Seven universal basic emotions of "sad", "surprise", "fear", "angry", "neutral", "disgust", and "happy" are considered as facial expression labels. In addition to these basic emotions, we asked annotators to choose all the emotions that may apply to an image. This resulted in secondary and tertiary dyads of seven basic emotions such as fear\_sad, fear\_anger, etc. Interestingly, this did not result in having combinations of three basic emotions. Figure \ref{FE_relations} shows the corresponding graph of seven basic emotions, their primary, secondary, and tertiary dydes presented in FePh dataset. Seven basic emotions are shown by colored circles with the emotion labels written inside them. Other colored circles connecting each two basic emotions illustrate the secondary or tertiary dyads of the basic emotions that are connected to. Emotion "Happy" has only one dyad, which is with "Neutral" emotion, named as "Neutral\_Happy". 

Although the FePh dataset presents annotated facial expression for all hand shape classes of the RWTH-PHOENIX-Weather 2014, we analyzed the facial expression labels for the top 14 hand shapes (i.e., classes "1", "index", "5", "f", "2", "ital", "b", "3", "b\_thumb", "s", "pincet", "a", "h", and "ae"). This is due to the demonstrated distribution of the counts per hand shape classes in \cite{cite8} that shows the top $14$ hand shape classes represent $90\%$ of the data. Seven universal basic facial expressions and their secondary or tertiary dyads occur with different frequencies in the data. Figure \ref{FE_frequency} shows the distribution counts per facial expression class in the data. As it shows, about $90\%$ of the data is expressed with basic facial expressions. In addition, Figures \ref{none_dist} and \ref{obscured_dist} illustrate the frequency of images with obscured faces and the "None" class in the top 14 hand shape classes, respectively.

By analyzing facial expressions per hand shape class, we found out that more than one facial expression class represents each hand shape class. Figure \ref{heatmaps} shows the frequency heatmaps of the seven facial expressions and their primary, secondary, and tertiary dyads for the top 14 hand shape classes. Each heatmap illustrates the frequency of facial expressions based on the facial expression graph of seven basic universal emotions and their primary, secondary, and tertiary dyads (shown in Figure \ref{FE_relations}) for one of the top 14 hand shape classes. The heatmaps show that more than one facial expression is expressed within a single hand shape class, which is due to the complexity of sign language in using facial expressions with hand shapes. Two of these complexities that affect performing different facial expressions within each hand shape class are as follows:

\begin{figure*}[t]
    \makebox[\linewidth][c]{%
    \centering
    \begin{subfigure}{.25\textwidth}
      \centering
      \includegraphics[width=\linewidth]{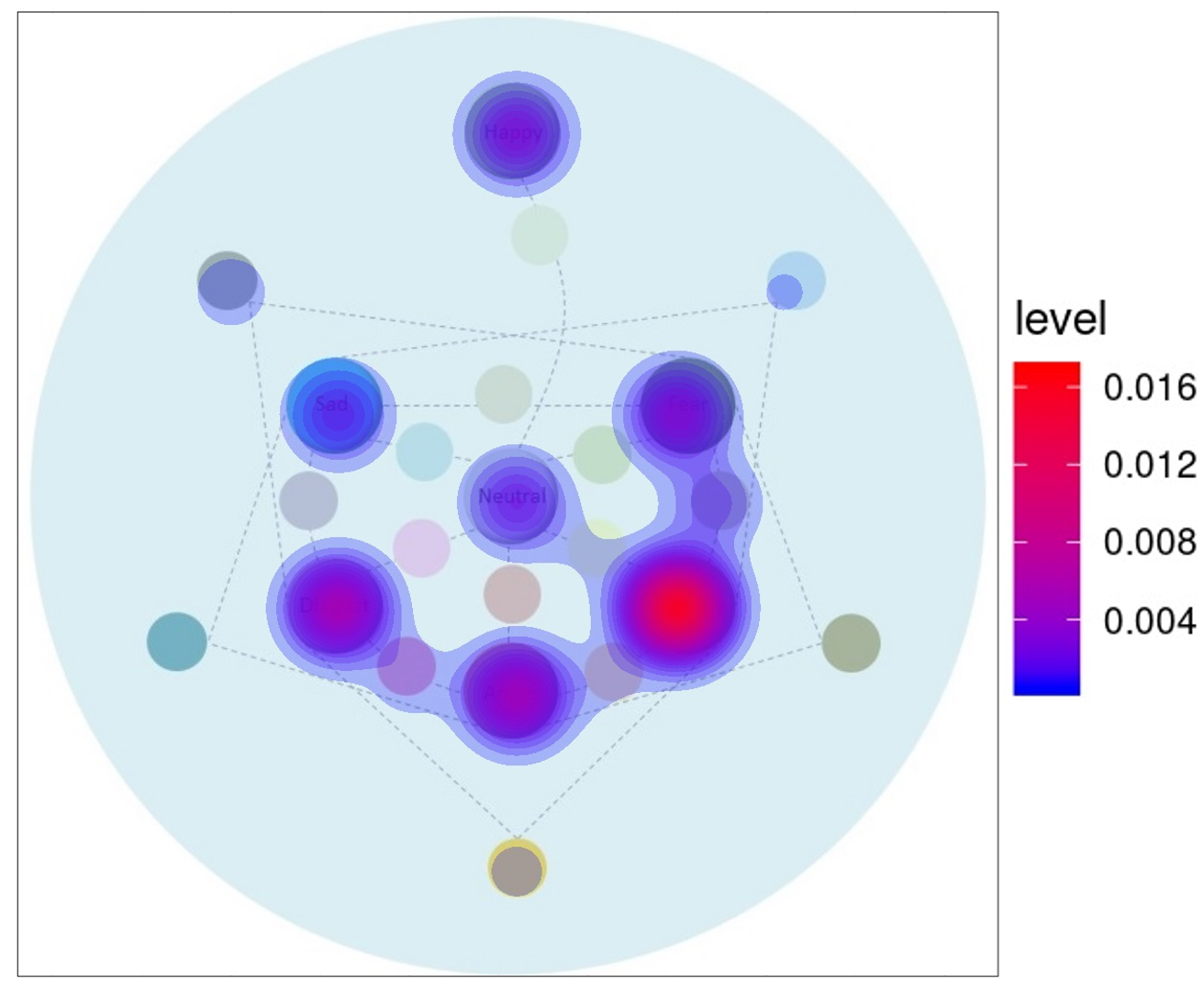}
      \caption{HSH\_1}
      \label{HSH_1}
    \end{subfigure}
    \begin{subfigure}{.25\textwidth}
      \centering
      \includegraphics[width=\linewidth]{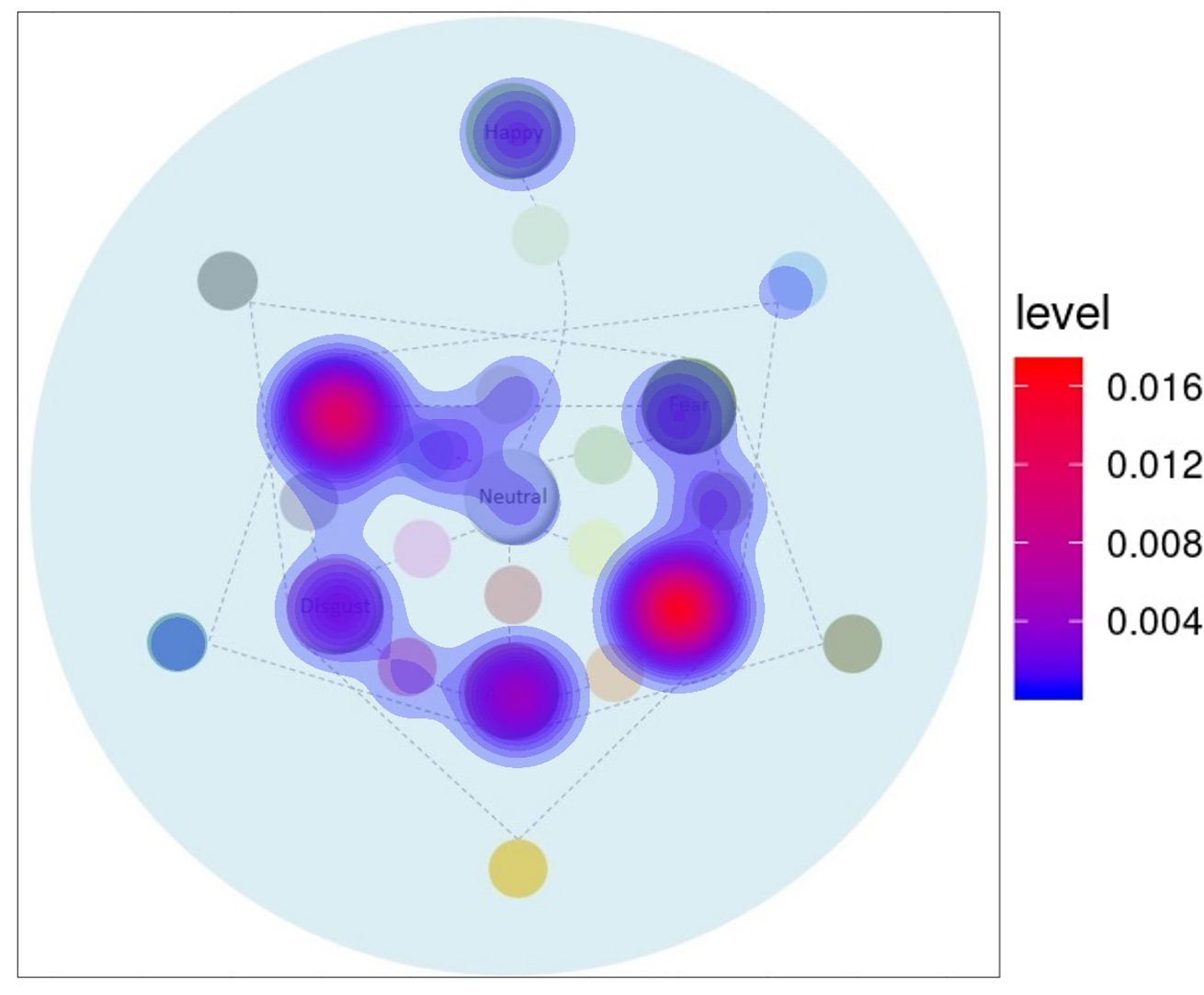}
      \caption{HSH\_index}
      \label{HSH_index}
    \end{subfigure}
    \begin{subfigure}{.25\textwidth}
      \centering
      \includegraphics[width=\linewidth]{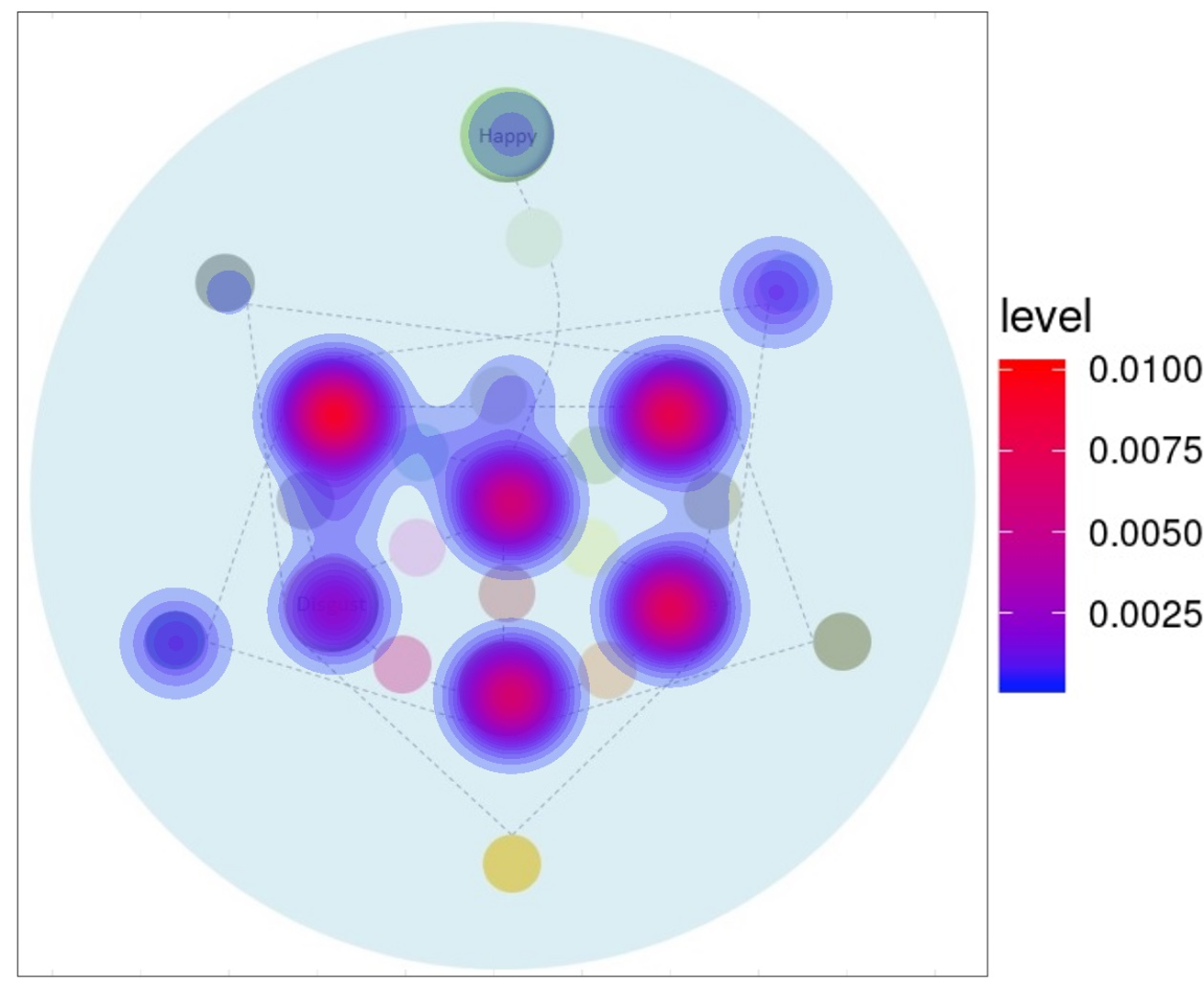}
      \caption{HSH\_5}
      \label{HSH_5}
    \end{subfigure}
    \begin{subfigure}{.25\textwidth}
      \centering
      \includegraphics[width=\linewidth]{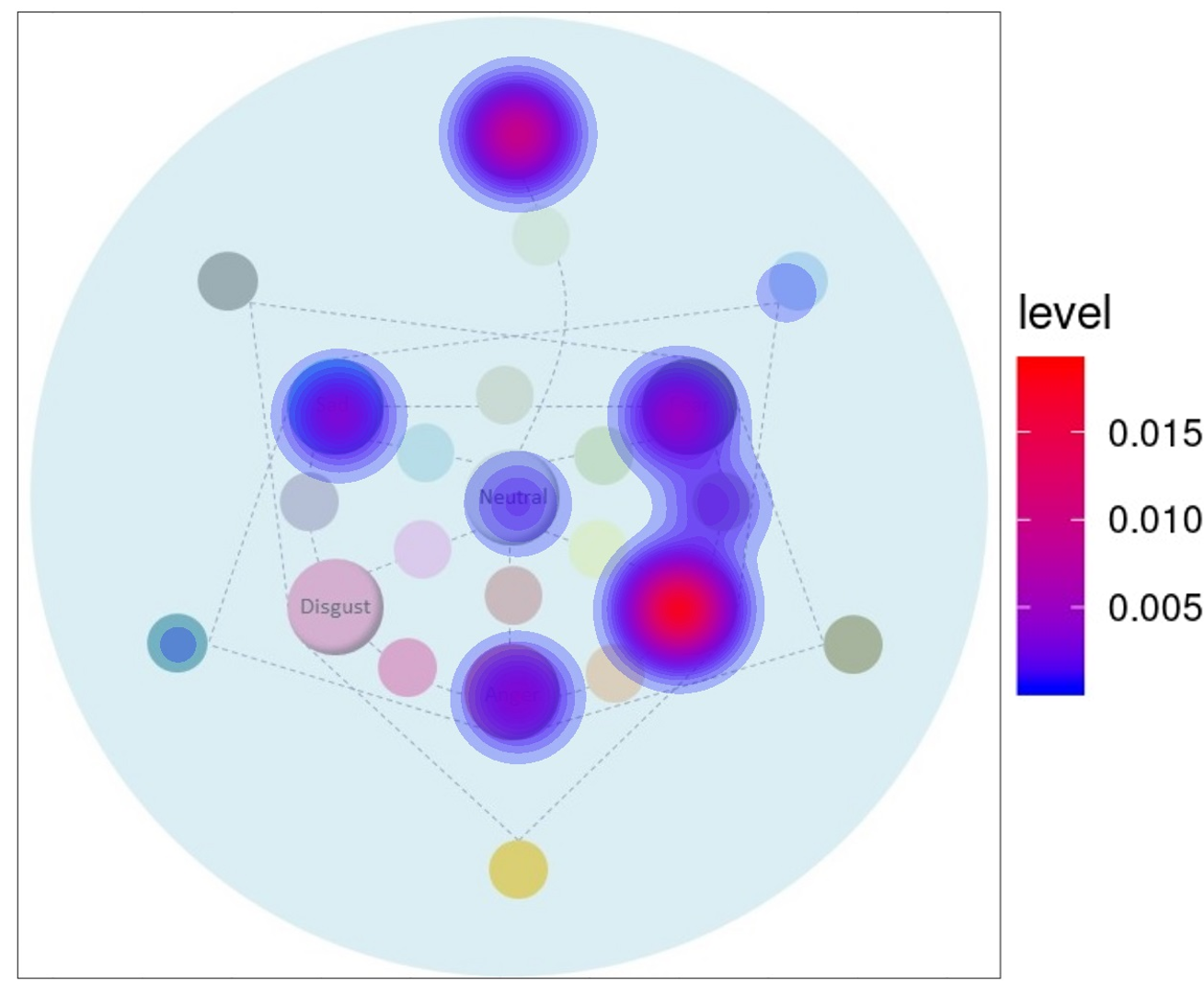}
      \caption{HSH\_f}
      \label{HSH_f}
    \end{subfigure}
    }
    
    \medskip
    
    \makebox[\linewidth][c]{%
    \centering
    \begin{subfigure}{.25\textwidth}
      \centering
      \includegraphics[width=\linewidth]{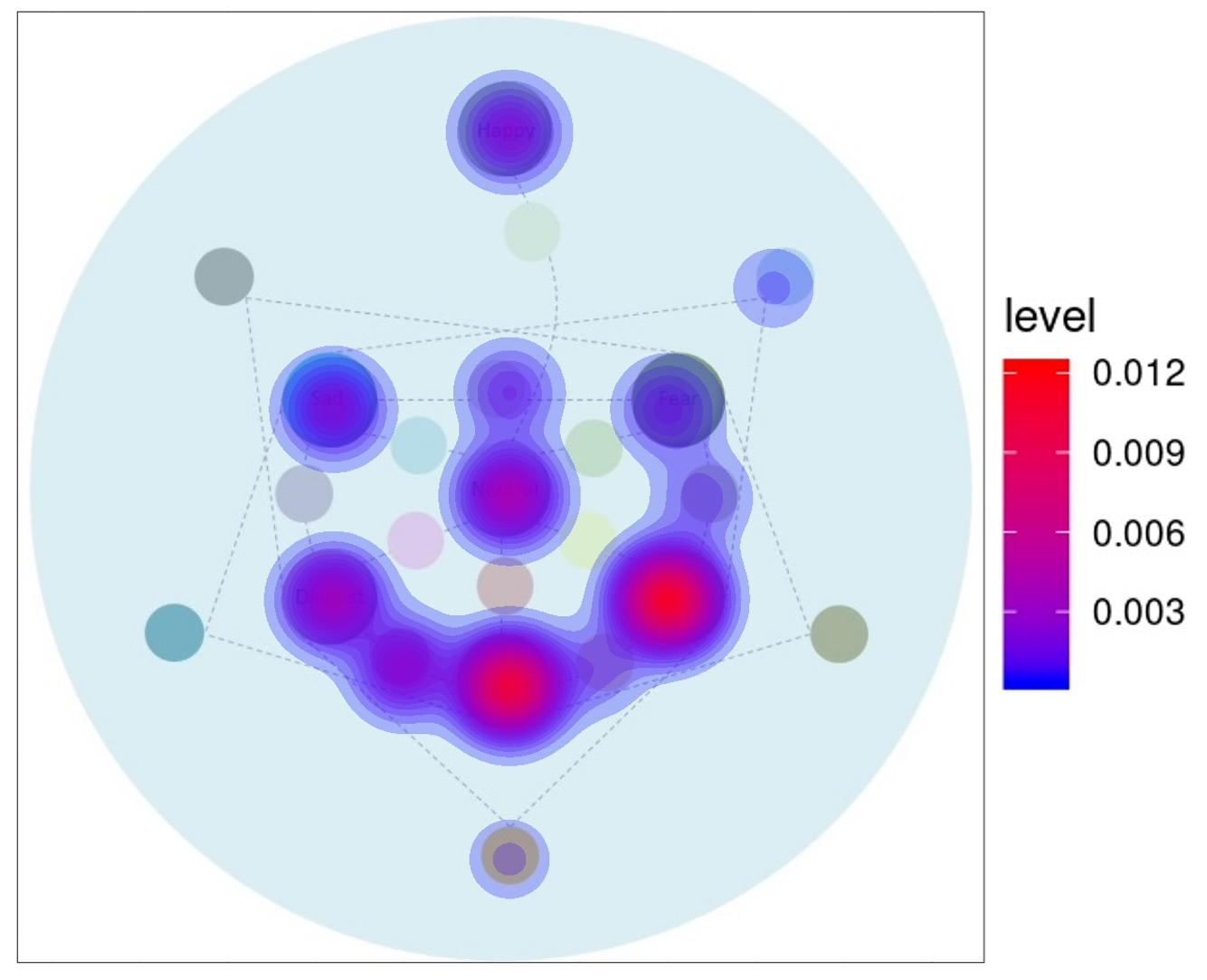}
      \caption{HSH\_2}
      \label{HSH_2}
    \end{subfigure}\hspace*{\fill}
    \begin{subfigure}{.25\textwidth}
      \centering
      \includegraphics[width=\linewidth]{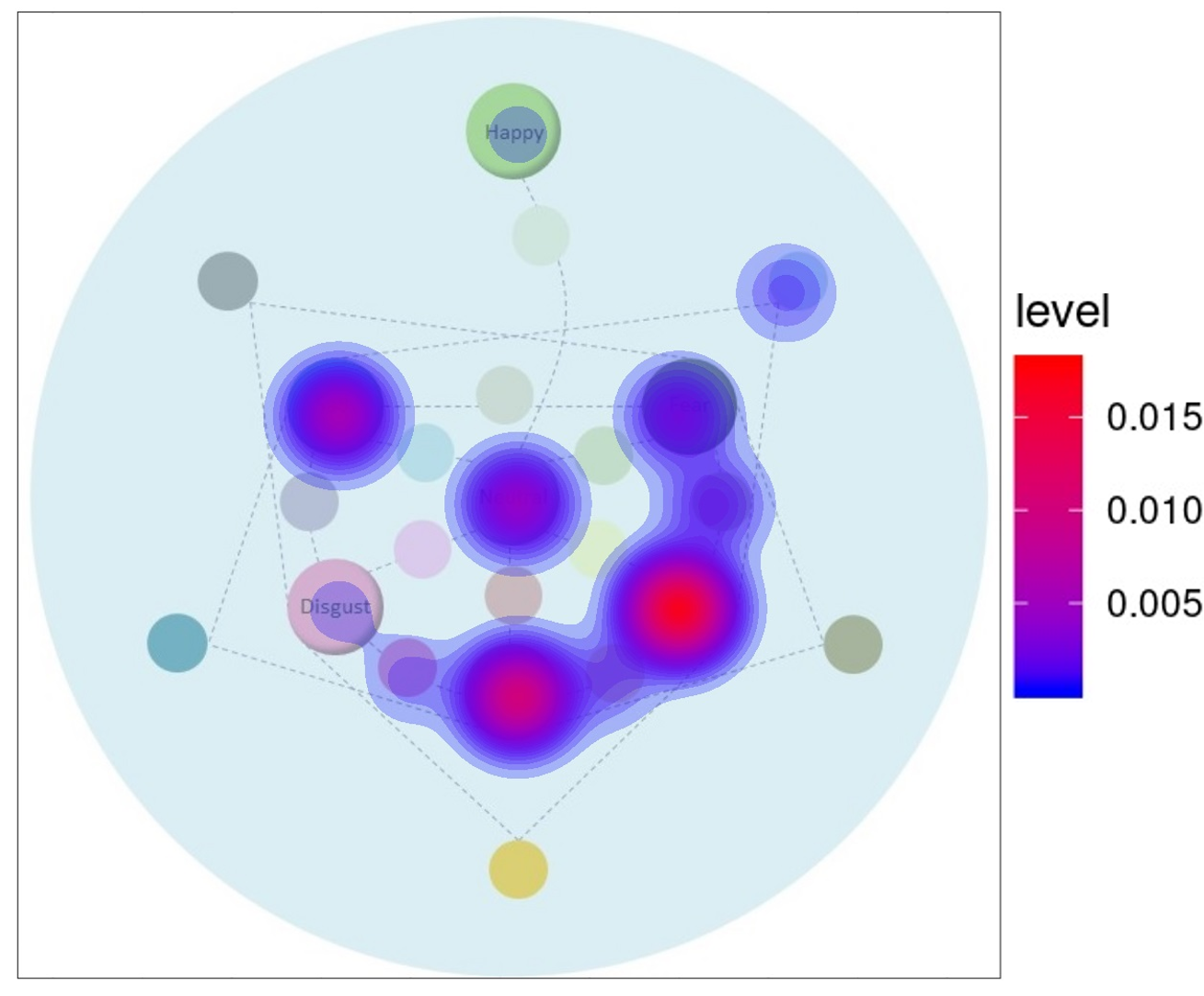}
      \caption{HSH\_ital}
      \label{HSH_ital}
    \end{subfigure}\hspace*{\fill}
    \begin{subfigure}{.25\textwidth}
      \centering
      \includegraphics[width=\linewidth]{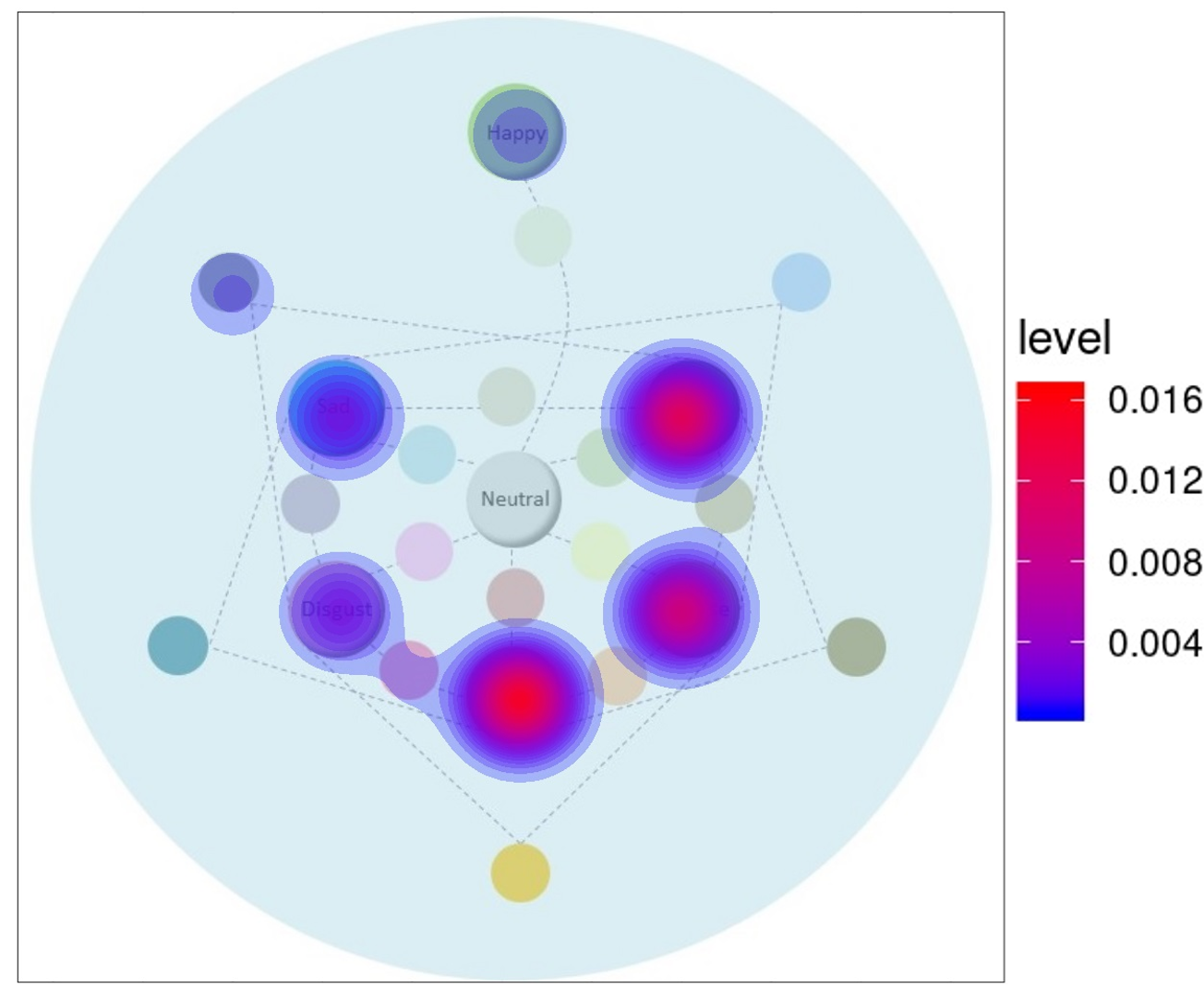}
      \caption{HSH\_b}
      \label{HSH_b}
    \end{subfigure}\hspace*{\fill}
    \begin{subfigure}{.25\textwidth}
      \centering
      \includegraphics[width=\linewidth]{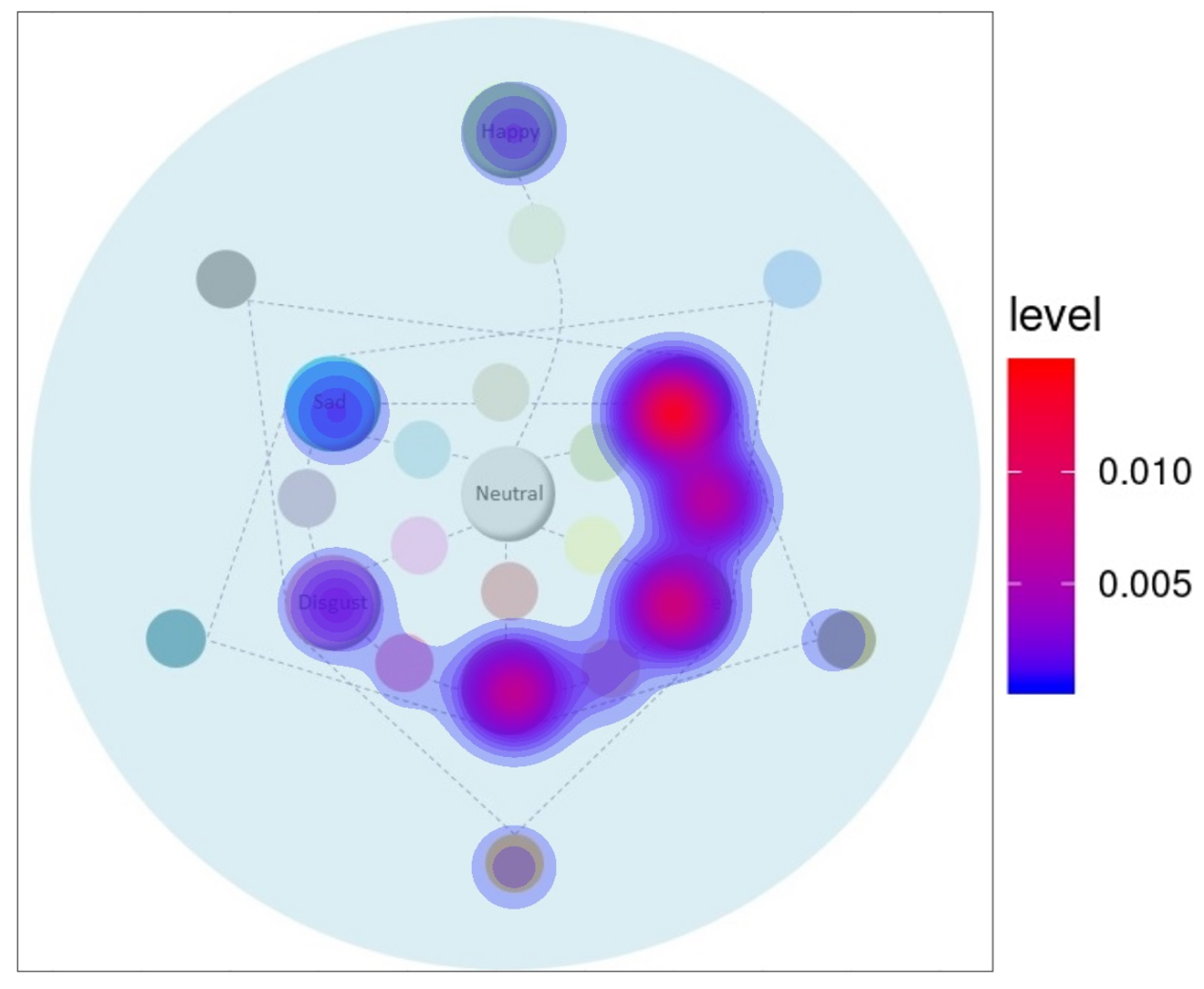}
      \caption{HSH\_3}
      \label{HSH_3}
    \end{subfigure}\hspace*{\fill}
    }
    
    \makebox[\linewidth][c]{%
    \begin{subfigure}{.25\textwidth}
      \centering
      \includegraphics[width=\linewidth]{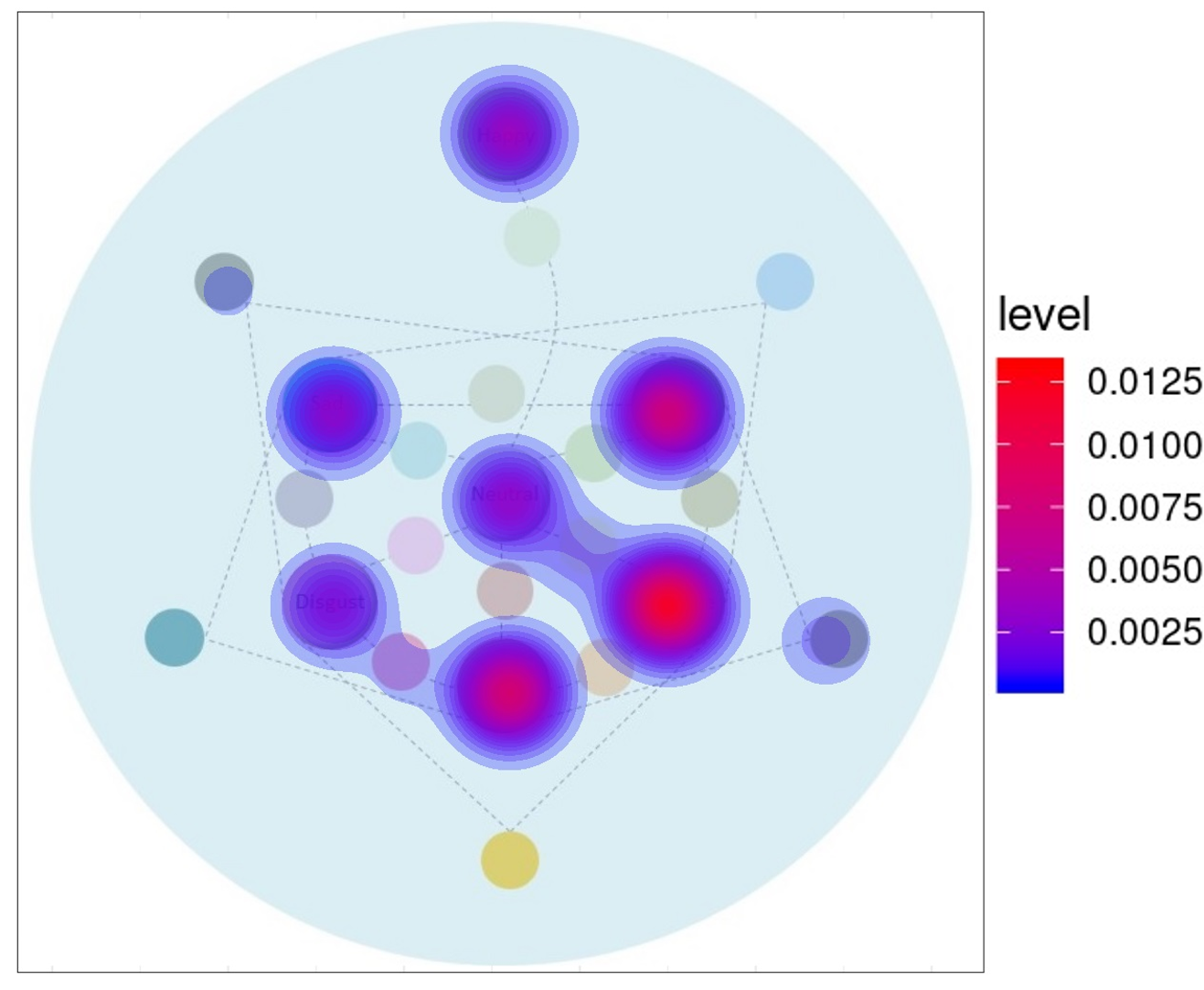}
      \caption{HSH\_b\_thumb}
      \label{HSH_b_thumb}
    \end{subfigure}\hspace*{\fill}
    \begin{subfigure}{.25\textwidth}
      \centering
      \includegraphics[width=\linewidth]{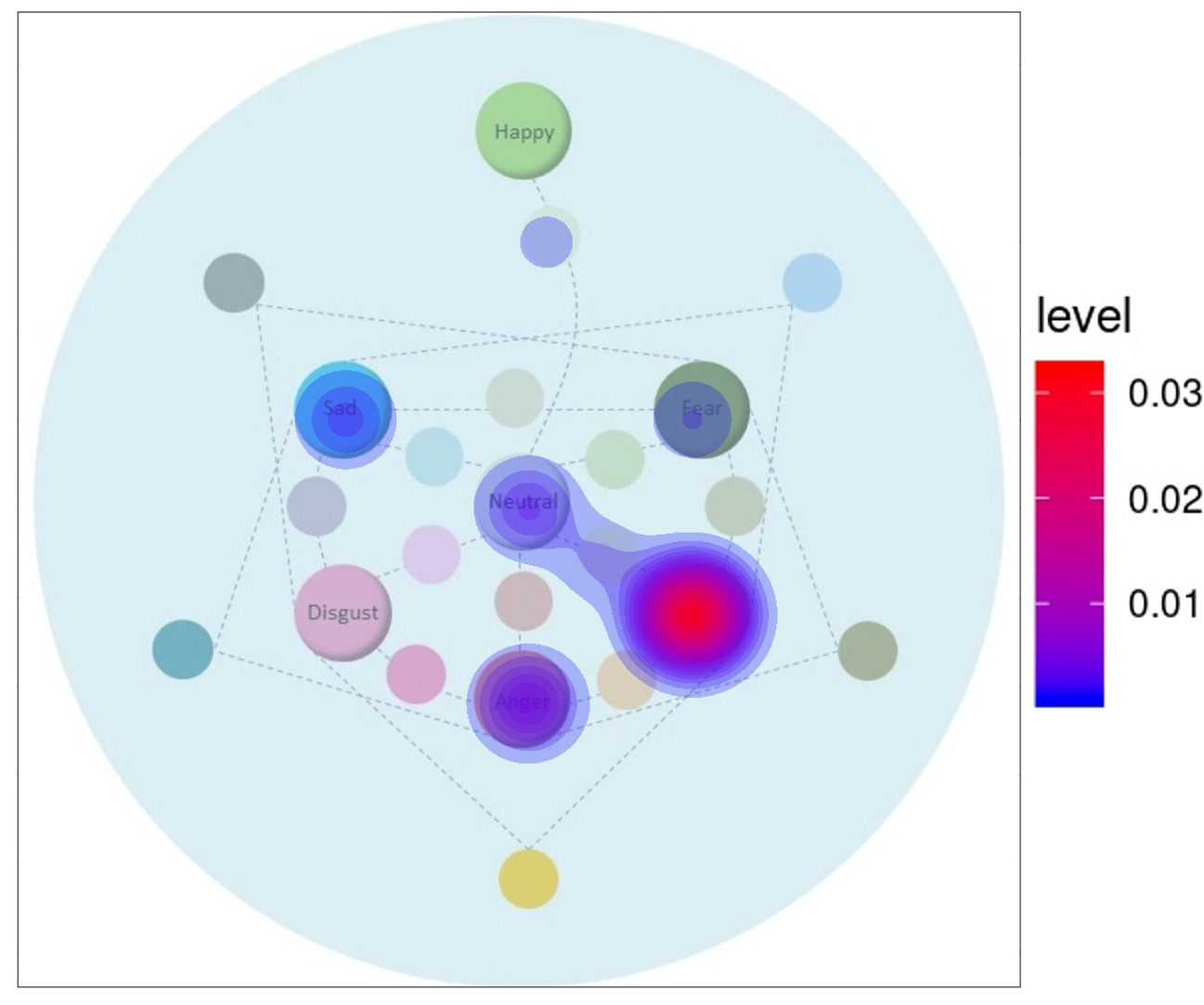}
      \caption{HSH\_s}
      \label{HSH_S}
    \end{subfigure}\hspace*{\fill}
    \begin{subfigure}{.25\textwidth}
      \centering
      \includegraphics[width=\linewidth]{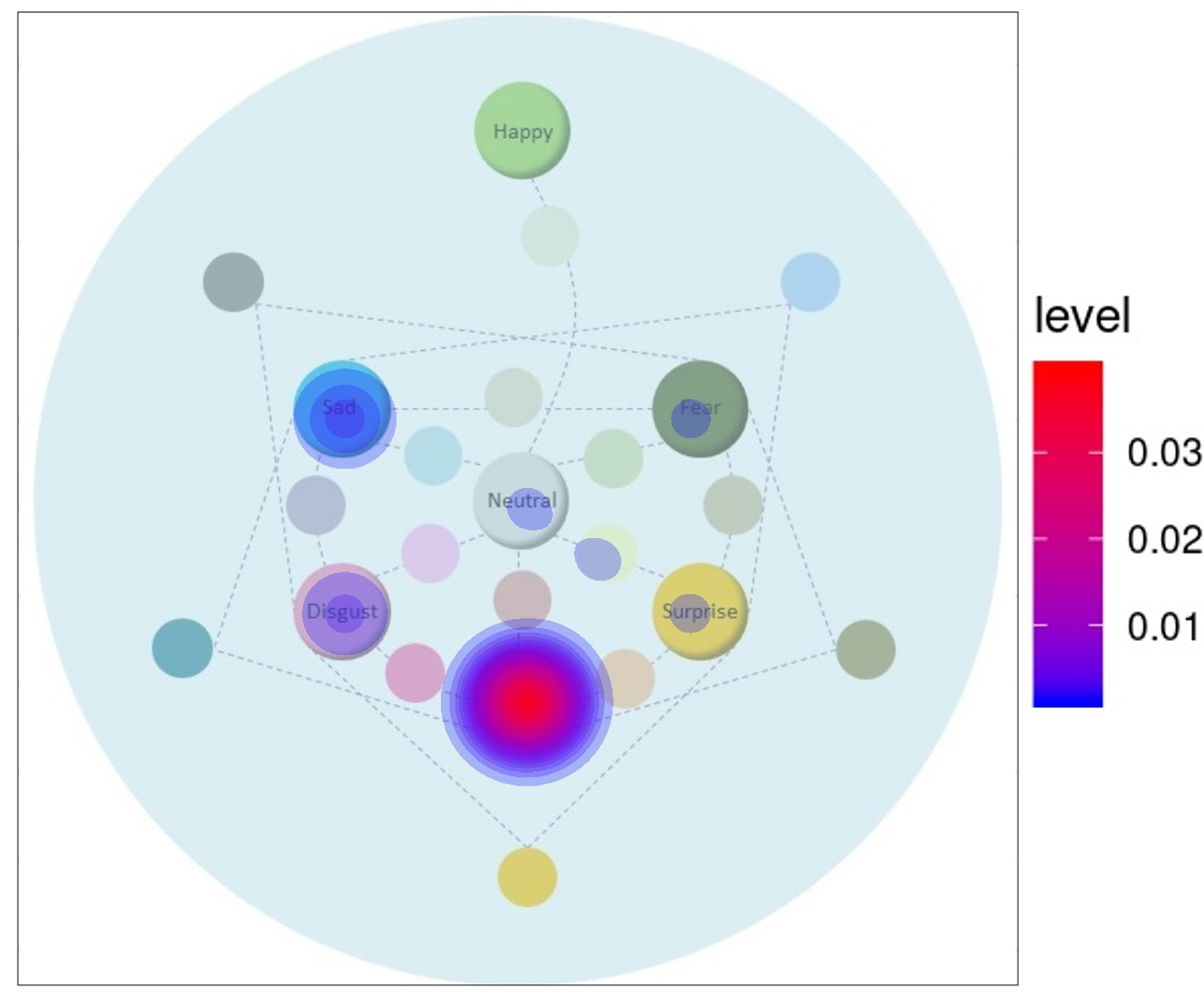}
      \caption{HSH\_pincet}
      \label{HSH_pincet}
    \end{subfigure}\hspace*{\fill}
    \begin{subfigure}{.25\textwidth}
      \centering
      \includegraphics[width=\linewidth]{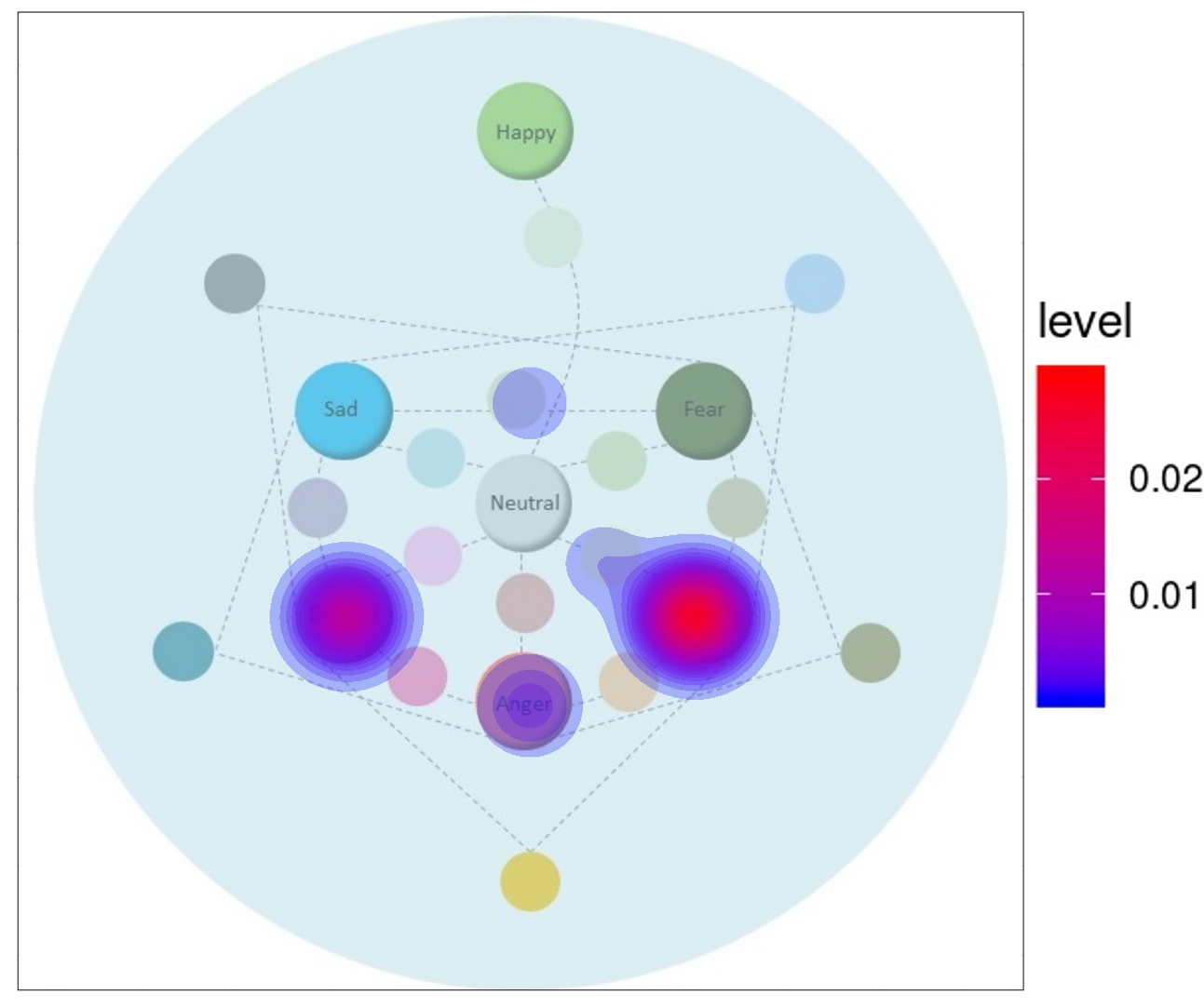}
      \caption{HSH\_a}
      \label{HSH_a}
    \end{subfigure}\hspace*{\fill}
    }
    
    \makebox[\linewidth][c]{%
    \centering
    \begin{subfigure}{.25\textwidth}
      \centering
      \includegraphics[width=\linewidth]{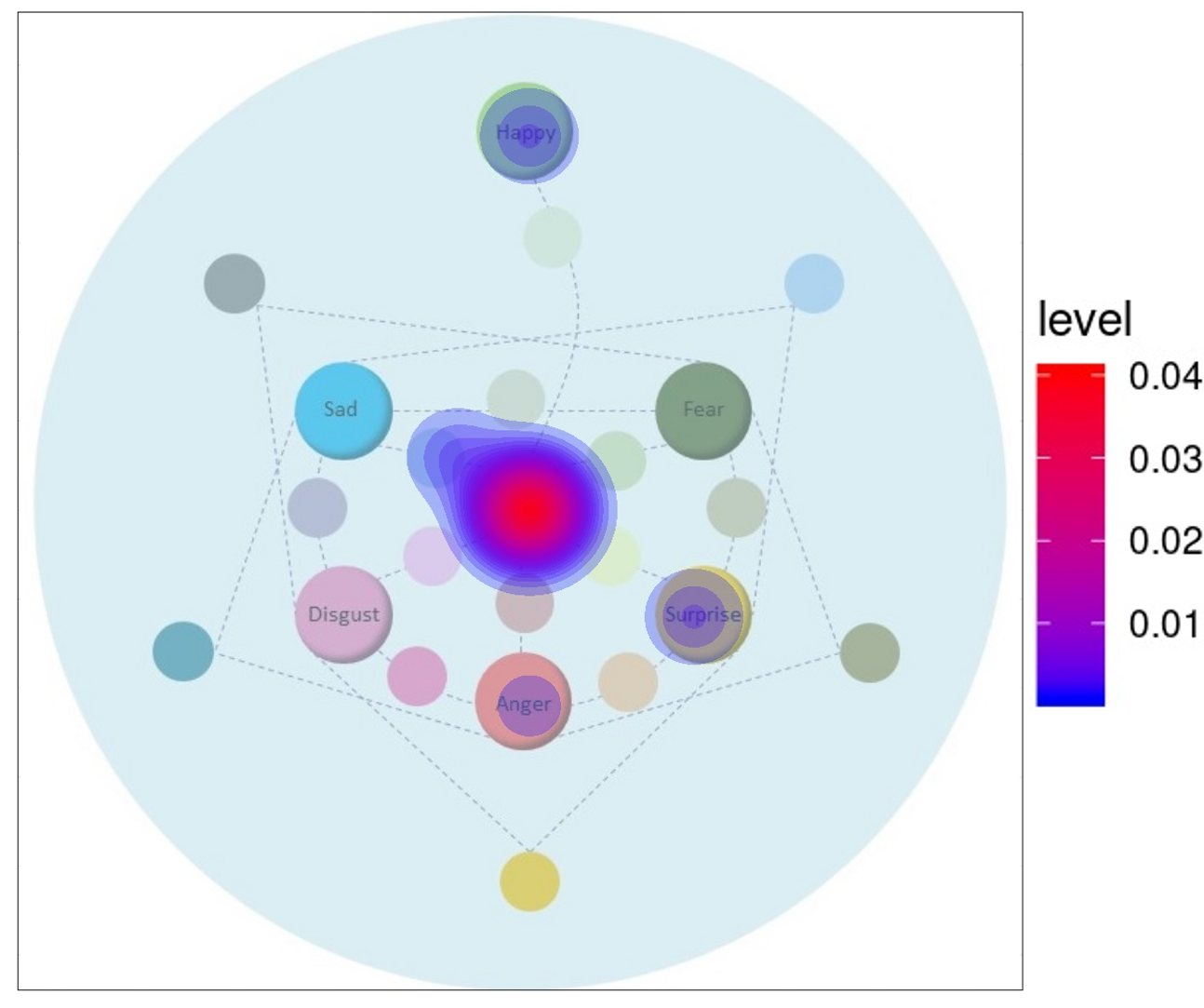}
      \caption{HSH\_h}
      \label{HSH_h}
    \end{subfigure}
    \begin{subfigure}{.25\textwidth}
      \centering
      \includegraphics[width=\linewidth]{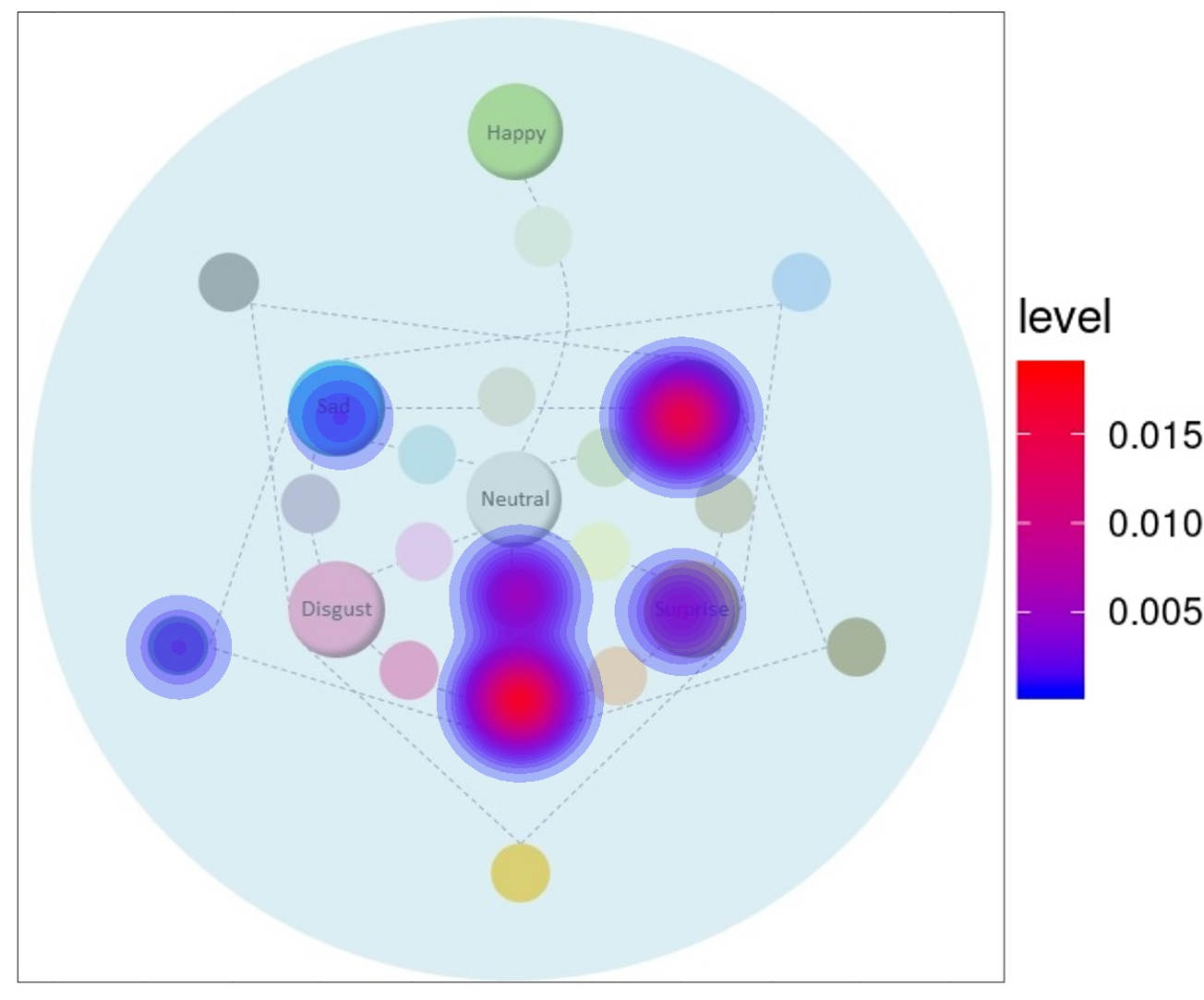}
      \caption{HSH\_ae}
      \label{HSH_ae}
    \end{subfigure}
    }
    \caption{Heatmaps showing the frequency distribution of the top 14 hand shape classes. Each heatmap, assigned to one hand shape class (briefly mentioned as HSH), shows the frequency of facial expressions on the assigned hand shape class over the facial expression graph of seven basic universal emotions and their primary, secondary, and tertiary dyads. As they show, more than one facial expressions are expressed within a single hand shape class.}
    \label{heatmaps}
\end{figure*}

\begin{figure*}[t]
\begin{subfigure}{.6\textwidth}
  \centering
  \includegraphics[width=0.9\linewidth]{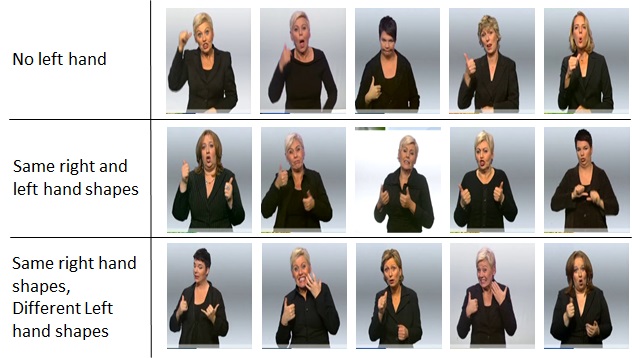}
  \caption{Exemplary images of hand shape class "1"}
  \label{intraclass}
\end{subfigure}\hspace*{\fill}
\begin{subfigure}{.4\textwidth}
  \centering
  \includegraphics[width=0.6\linewidth]{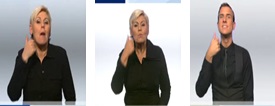}
  \caption{Full frame images with similar hand shapes but different facial expressions}
  \label{similarhands}
\end{subfigure}
\caption{Exemplary full frame images of the RWTH-PHOENIX-Weather 2014 dataset}
\label{exampleimage}
\end{figure*}

\normalsize
\begin{table*}[t]
\captionsetup{width=1\textwidth}
\caption{The correlation matrix of facial expressions' dummy variables and their frequencies for the top 14 hand shape classes. Since only the correlation between facial emotions and their frequencies in each hand shape\\ class are important here, the first column of facial expressions and their frequencies for each hand shape class is presented. Empty cells show the absence of the correlation (the facial expression is not expressed in the hand shape class images).}
\label{correlation }
\centerline{
\resizebox{1\textwidth}{!}{
\begin{tabular}{l|cccccccccccccc}
 & \multicolumn{14}{c}{Frequency of facial expressions for hand shape class} \\
 \hline
 & 1 & index & 5 & f & 2 & ital & b & 3 & b\_thumb & s & pincet & a & h & ae \\[5pt]
  \hline
Frequency & 1 & 1 & 1 & 1 & 1 & 1 & 1 & 1 & 1 & 1 & 1 & 1 & 1 & 1 \\
neutral & -0.027 & -0.145 & 0.153 & -0.210 & 0.115 & 0.066 & 0.666 & -0.221 & -0.019 & -0.168 & -0.150 & -0.371 & 0.995 &  \\
anger & 0.194 & 0.114 & 0.176 & -0.047 & 0.582 & 0.405 & -0.164 & 0.287 & 0.343 & -0.093 & 0.991 & 0.087 & -0.259 & 0.350 \\
disgust & 0.233 & -0.024 & -0.067 & 0.041 & 0.075 & -0.231 & 0.459 & -0.076 & -0.099 & -0.242 & -0.112 &  &  &  \\
fear & 0.064 & -0.089 & 0.257 & 0.335 & -0.126 & -0.061 & -0.277 & 0.697 & 0.303 & -0.192 & -0.150 &  &  & 0.250 \\
happy & 0.038 & -0.080 & -0.206 & -0.036 & -0.045 & -0.241 & -0.126 & -0.124 & 0.022 & 0.850 & -0.073 &  & -0.209 &  \\
sad & -0.040 & 0.502 & 0.338 & 0.805 & -0.032 & 0.130 & 0.308 & -0.124 & -0.019 & 0.379 & -0.170 &  &  & -0.350 \\
surprise & 0.857 & 0.737 & 0.234 & 0.139 & 0.662 & 0.787 & -0.164 & 0.383 & 0.624 &  & -0.015 & 0.612 & -0.209 & -0.250 \\
none & -0.053 & 0.187 & 0.651 &  & -0.005 & 0.034 &  & 0.093 & 0.263 &  &  & 0.546 & -0.109 & 0.751 \\
anger\_neutral &  &  &  &  &  &  &  &  &  &  &  &  &  & -0.150 \\
disgust\_anger & -0.170 & -0.153 &  &  & -0.005 & -0.199 & -0.353 & -0.196 & -0.300 &  &  &  &  &  \\
anger\_fear &  &  &  &  &  &  &  & -0.196 & -0.260 &  &  &  &  &  \\
fear\_disgust & -0.170 &  & -0.230 &  &  &  & -0.296 &  & -0.300 &  &  &  &  &  \\
happy\_neutral &  &  &  &  &  &  &  &  &  & -0.292 &  &  &  &  \\
neutral\_sad &  & -0.097 & -0.218 &  &  &  &  &  &  &  &  &  & -0.209 &  \\
sad\_anger &  & -0.169 & -0.160 & -0.298 &  &  &  & -0.221 & -0.300 &  &  &  &  & -0.350 \\
sad\_disgust &  & -0.169 & -0.195 &  &  &  &  &  &  &  &  &  &  &  \\
sad\_fear &  & -0.145 & -0.206 &  & -0.139 &  &  &  &  &  &  & -0.437 &  &  \\
neutral\_surprise & -0.222 &  &  &  &  &  &  &  & -0.260 & -0.242 & -0.150 & -0.437 &  &  \\
surprise\_anger & -0.209 &  &  &  & -0.192 & -0.178 &  & -0.148 &  &  &  &  &  &  \\
surprise\_disgust & -0.183 &  &  &  & -0.219 &  &  & -0.172 &  &  &  &  &  &  \\
surprise\_fear & -0.118 & -0.113 & -0.241 & -0.189 & -0.192 & -0.125 &  & 0.214 &  &  &  &  &  &  \\
surprise\_sad & -0.190 & -0.169 &  & -0.287 & -0.219 & -0.178 &  &  &  &  &  &  &  &  \\
\end{tabular}%
}}
\end{table*}

First, although some meanings are communicated using only one hand (usually the right hand), many sign language meanings are communicated using both hands with different hand poses, orientations, and movements.
Figure \ref{intraclass} shows some exemplary full frame images of hand shape class "1" of the RWTH-PHOENIX-Weather 2014 dataset with different facial expressions. As the figure illustrates, the usage of right hand shapes have large intra-class variance (i.e., the left hand may not be used or may perform similar or different hand shape from the right hand) that may affect the meanings, and as a result, the facial expressions corresponding to them. The first top row in the figure shows the full frame images with the right hand shape of class "1" and no left hand shape. The second row shows some exemplary full frame images, in which the signer has used both hands. In this row, although both right and left hands demonstrate the same hand shape class (hand shape class "1"), their pose, orientation, and movement can differ, which may affect the corresponding facial emotion. The third row of images in Figure \ref{intraclass} illustrates full frame examples of using both hands with different hand shapes and facial expressions. Therefore, although RWTH-PHOENIX-Weather 2014 MS Handshapes is a valuable resource presenting right hand shape labels, it lacks pose, orientation, and movement labels of the right hand along with the left hand shape labels. Adding this information to the data affects the communicated meanings as well as the facial expressions that are expressed.

Second, due to the communication of grammar via facial expressions, identical hand shapes may be performed with different facial expressions. Figure \ref{similarhands} demonstrates some images of this kind that despite the similarity of hand shapes, the facial expressions are different. This complex usage of hands with large intra-class variance and inter-class similarities help signers to communicate different meanings with similar or different facial expressions. 

In addition to the above, the frequency of facial expressions expressed in each hand shape class shows evidence of a meaningful association between hand shapes and facial expressions in the data. To better illustrate this correlation, we calculated the correlation matrix of facial expressions' frequency in each hand shape class. Since the correlations between the facial expressions together is not the focus of this manuscript, the first column of each correlation matrix that shows the correlation between frequency and each facial expression is only considered. Table \ref{correlation } illustrates the first columns of facial expressions and their frequencies of occurrence in the top 14 hand shape classes correlation matrices. Monitoring each column in Table \ref{correlation } gives the most correlated facial expressions for each hand shape. For example, in the column of hand shape class "3", the positive values of 0.697, 0.383, 0.287, 0.214 and 0.093 (that are in intersections with "fear", "surprise", "anger", "surprise\_fear" and "None" respectively) show the positive correlation values with facial expressions in hand shape class "3". These highly correlated facial expressions in each hand shape class can also be interpreted from heatmaps illustrated in Figure \ref{heatmaps}. For example, the heatmap of hand shape class "3" (shown in Figure \ref{HSH_3}) indicates that for signers signing hand shape class "3", the distribution of the facial expression label counts is weighted towards expressing more of the "fear" emotion, which has the highest correlation value in the hand shape class "3".

\section{Usage Notes}
\IEEEPARstart{T}{o} the best of our knowledge, this dataset is the first annotated vision-based publicly available sequenced facial expression dataset in the context of sign language. Although the number of facial images is enough for statistical and some machine learning methods, it may not be sufficient for some of the state-of-the-art learning techniques in the field of computer vision. Therefore, for such studies, we suggest users to create matched samples choosing subjects from the dataset. 
This work not only provides an annotated facial expression dataset with different head poses, orientations, and movements, but also contributes in availability of a sign language dataset with both hand shape and facial expression labels with attributions in multi-modal future works in the field. In addition, this dataset has a wider application in other research areas such as gesture recognition and Human-Computer Interaction (HCI) systems.

\section{Conclusion}\label{conclusion}
In this work, we presented the FePh dataset, which to the best of our knowledge, is the first real-life annotated sequenced facial expression dataset in the context of sign language. FePh in conjunction with RWTH-PHOENIX-Weather 2014 and RWTH-PHOENIX-Weather 2014 MS Handshapes datasets constitute the first sign language data with both handshapes and facial expression labels. We hope this unique characteristic will propel research in multi-modal sign language and gesture recognition.

As preliminary results and analysis of the data indicate a meaningful relationship between two important modals of sign language (i.e., handshapes and facial expressions), for future work, we propose applying multi-modal learning and computer vision techniques on joint RWTH-PHOENIX-Weather 2014 MS Handshapes and FePh datasets. We believe that the introduction of this dataset will allow the facial expression, sign language, and gesture recognition communities to improve their learning techniques to the latest levels of computer vision trends.

\section*{Acknowledgment}

The authors would like to thank Hanin Alhaddad, Nisha Baral, Kayla Brown, Erdais Comete, Sara Hejazi, Jasser Jasser, Aminollah Khormali, Toktam Oghaz, Amirarsalan Rajabi, Mostafa Saeidi, and Milad Talebzadeh for their assistance in annotating the data.

\ifCLASSOPTIONcaptionsoff
  \newpage
\fi



%

%


\begin{IEEEbiographynophoto}{Marie Alaghband}
is a qualified PhD student of Industrial Engineering department and a member of the Complex Adaptive Systems Laboratory (CASL) at the University of Central Florida (UCF). Marie received her M.Sc. in Socio-economic systems Engineering (The Institute for Management and Planning Studies; 2013-2016), and a B.Sc. in Statistics (Isfahan University of Technology; 2007-2012). She is currently a graduate teaching and research assistant in the Industrial Engineering and Management Systems (IEMS) department at UCF. Her research expertise in her master’s degree lies in scheduling, operation research, and economics, while in her PhD are machine learning and data analysis in Industrial engineering related topics.

Research Interest:
Machine Learning, Computer Vision, Deaf Education and Study, Data Analysis
\end{IEEEbiographynophoto}


\begin{IEEEbiographynophoto}{Niloofar Yousefi} is the science director for the UCF Complex Adaptive Systems (CASL) Lab and a post-doctoral research associate in Machine Learning and Computational Learning Theory. She received her Ph.D. in Industrial Engineering/Operation Research with an emphasis in Computer Science, Machine Learning, and Pattern Recognition. She has a Master of Science in Operation Research from University of Tehran and a Bachelor of Science in Applied Mathematics from Iran University of Science and Technology.

Research Interest:
Machine Learning, Statistical Learning Theory, Pattern Recognition, Kernel-based Models, Multi-Task Learning
\end{IEEEbiographynophoto}


\begin{IEEEbiographynophoto}{Ivan Garibay}
serves as the director of the Complex Adaptive Systems Laboratory (CASL) and the Master of Science in Data Analytics (MSDA) program at the University of Central Florida (UCF). Dr. Garibay received his B.Sc. and Diploma in EE (Ricardo Palma University; 1994,1995), and MS, Ph.D. in CS (University of Central Florida; 2000, 2004). He is currently an Assistant Professor in the Industrial Engineering and Management Systems (IEMS) department at UCF. His research expertise lies in complex systems, agent-based models, data and network science, artificial intelligence and machine learning. His research is currently sponsored by federal agencies and industry by more than $7.0$ M. He has published and presented more than 75 papers in journals and conferences.

Research Interest:
Complex systems, Computational Social Science, Artificial Intelligence, Innovation
\end{IEEEbiographynophoto}




\end{document}